\PassOptionsToPackage{table}{xcolor}
\documentclass[fleqn,10pt]{wlscirep}
\usepackage[utf8]{inputenc}
\usepackage[T1]{fontenc}
\usepackage{multirow}
\usepackage{graphicx}
\usepackage{algorithm}
\usepackage{float}
\usepackage{algpseudocode}
\PassOptionsToPackage{table}{xcolor}
\usepackage[table]{xcolor}

\usepackage{color,soul}

\title{ \includegraphics[height=1.0cm]{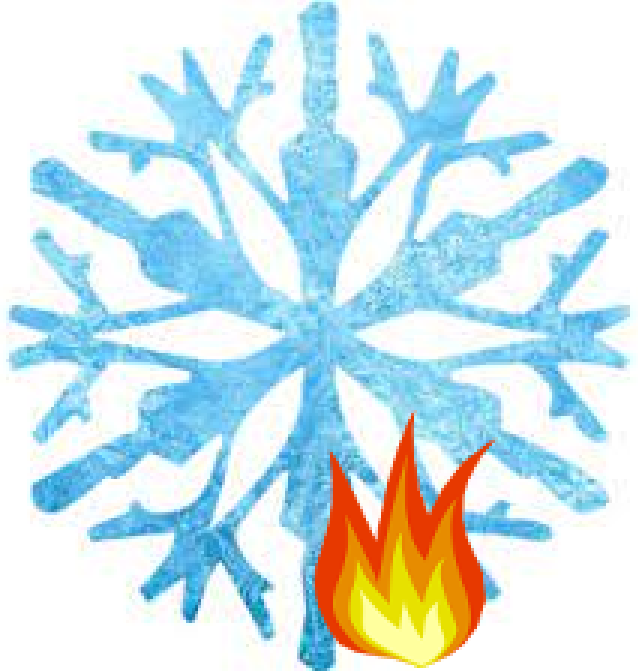} Parameter-Efficient Fine-Tuning of Large Language Models using Semantic Knowledge Tuning}

\author[1]{Nusrat Jahan Prottasha}
\author[2,+]{Asif Mahmud}
\author[3,+]{Md. Shohanur Islam Sobuj}
\author[4]{Prakash Bhat}
\author[1]{Md Kowsher}
\author[1]{Niloofar Yousefi}
\author[1]{Ozlem Ozmen Garibay}

\affil[1]{University of Central Florida
Orlando, 32816, Florida}
\affil[2]{Noakhali Science \& Technology University, Noakhali, 3814, Bangladesh}
\affil[3]{Hajee Mohammad Danesh Science \& Technology University, Dinajpur, 5200,  Bangladesh}
\affil[4]{Amazon, New Jersey, USA}

\affil[+]{these authors contributed equally to this work}

\begin{abstract} Large Language Models (LLMs) are gaining significant popularity in recent years for specialized tasks using prompts due to their low computational cost. Standard methods like prefix tuning utilize special, modifiable tokens that lack semantic meaning and require extensive training for best performance, often falling short. In this context, we propose a novel method called Semantic Knowledge Tuning (SK-Tuning) for prompt and prefix tuning that employs meaningful words instead of random tokens. This method involves using a fixed LLM to understand and process the semantic content of the prompt through zero-shot capabilities. Following this, it integrates the processed prompt with the input text to improve the model's performance on particular tasks. Our experimental results show that SK-Tuning exhibits faster training times, fewer parameters, and superior performance on tasks such as text classification and understanding compared to other tuning methods. This approach offers a promising method for optimizing the efficiency and effectiveness of LLMs in processing language tasks.

\end{abstract}
\begin{document}

\flushbottom
\maketitle
%
%
\thispagestyle{empty}

\section{Introduction}

The domain of NLP has seen a remarkable transformation in recent years, primarily driven by the introduction of LLMs \cite{zhu2023large}. Transformer-based Language Models (TLMs) initially led about a revolution by showing outstanding capabilities in capturing extensive dependencies \cite{greco2023bringing}. However, the challenges connected with adapting TLMs to various tasks, combined with their resource-intensive training, resulted to the development of more powerful models, such as GPT-3 \cite{brown2020language}. With billions of parameters, these LLMs have not only boosted performance benchmarks across various tasks but have also extended their applications into novel domains, including creative writing and multimodal learning \cite{mcintosh2023google}.

Despite their notable achievements, a in-depth analysis of LLMs reveals several major limitations. The extensive computational resources required for their training raise questions about environmental sustainability and restrict accessibility to research facilities with sufficient equipment and resources \cite{raiaan2023review}.

To address this issue, there has been an increasing focus on the latest innovations in parameter-efficient fine-tuning methods\cite{liu2022fewshot}.As compared to retraining the entire model from scratch, fine-tuning LLMs has proven to be a more rapid and efficient approach. Nevertheless, the fine-tuning of all parameters of LLMs remains a challenge due to their vast size, which typically consists of billions of parameters. Despite this, the fine-tuning process still requires extensive computational resources, much like the pretraining technique.

Adapting to this challenge, adapter training has gained importance as a more efficient approach\cite{pmlr-v97-houlsby19a}. This approach involves introducing domain-specific parameters, referred to as adapters, into pretrained models. These adapters, which are made of small neural networks, are strategically inserted within or between the layers of the pretrained model. During the training process, only the parameters of these added adapters are updated, while the parameters of the pretrained model remain unchanged \cite{pmlr-v97-houlsby19a}.

While adapters provide a more simple approach, they may not fully capture complex data patterns as effectively as fine-tuning the entire model. In addition, determining the optimal locations to insert adapters within the LLM can be challenging and may require experimentation. Nonetheless, prompt tuning complements adapter training by offering additional contextual information to guide the model's understanding of the task at hand.

Prompt tuning\cite{lester2021power} does not modify the underlying structure of the model, potentially resulting in quicker inference times and decreased resource consumption in contrast to utilizing adapters. Furthermore, prefix tuning\cite{li2021prefixtuning} has been proposed as a method to improve performance. Unlike prompt tuning, which updates only a portion of a single layer, prefix tuning updates a section of every layer consistently, and it has shown improved performance in succeeding tasks.

The use of prompts and prefix tuning techniques\cite{lester2021power, li2021prefixtuning, kowsher2023tuning} can pose challenges in terms of the effectiveness and interpretability of the employed prompts or prefixes. These methods generally utilize trainable virtual tokens within an adapter, which may not have essential semantic significance and require extensive training to acquire domain-specific knowledge efficiently. Consequently, the performance of these techniques may not be optimal, particularly when dealing with complex tasks, and extensive training is necessary to achieve optimal performance.

To overcome these challenges, we propose SK-Tuning, a novel approach that focuses on improving the performance of fine-tuning LLMs for prompt and prefix tuning. Unlike standard techniques that depend on random virtual tokens, SK-Tuning utilizes genuine, semantically rich prompts or prefixes for adapter training. By employing the LLM's innate capacity to understand linguistic semantics, SK-Tuning strives to improve performance by integrating semantic knowledge directly from prompts or prefixes.

LLMs display remarkable zero-shot capabilities, allowing them to perform tasks without explicit training, as shown in recent studies \cite{huang2022language}. To maximize the potential of these capabilities, SK-Tuning utilizes LLM's ability to understand prompts or instructions in a zero-shot manner. This approach speeds up the convergence process during fine-tuning because we concentrate only on refining the semantic representation of the prompt or prefix.

\begin{figure}[ht]
    \centering
    \includegraphics[width=0.8\textwidth]{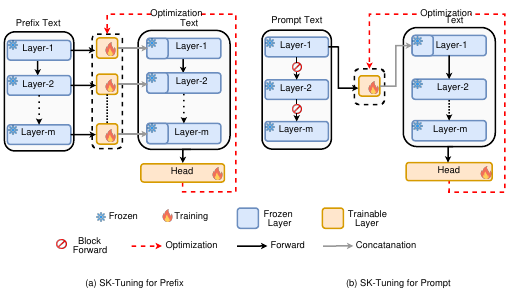}
    \caption{SK-Tuning approaches for Prefix (left) and Prompt (right). The \textcolor{red}{dashed line} represents the optimization path during the backward pass to the trainable adapter. Notably, in the context of prompt-tuning (on the right), the \textcolor{red}{no sign} signifies the discontinuation of the forward pass beyond a certain point. This is because we exclusively initialize layer-specific semantic information for the prompt, rendering the continuation of the forward pass unnecessary for the remaining layers. }
    \label{fig:sktuning}
\end{figure}

The SK-Tuning method is presented in Figure \ref{fig:sktuning}, which displays the stages for prompt and prefix tuning. At first, the entire LLM is frozen to maintain its pretrained knowledge. Next, the frozen LLM is utilized to extract the semantic representation from the prompt or prefix text. This representation is then educated with a small adapter to improve its task-specific intelligence. Lastly, the revised representation is combined with the embedding of the input text, guaranteeing that the model effectively integrates both the semantic context provided by the prompt or prefix and the textual data of the input.

We perform wide-ranging experimental evaluations across a variety of downstream tasks, including sequence classification, token classification, and NLI, to practically show the efficiency and excellence of SK-Tuning compared to traditional fine-tuning methods. Furthermore, we compare SK-Tuning with other parameter-efficient approaches, such as prompt tuning, prefix tuning, p-tuning, and LoRA, highlighting its unique advantages and contributions to the field of NLP.

The major contributions of this paper are
summarized as follows:

\begin{itemize}
    \item  This paper introduces SK-Tuning, a novel approach for fine-tuning LLMs using real, semantically meaningful prompt or prefix text.
    
    \item SK-Tuning improves training efficiency and convergence speed by utilizing the inherent semantic understanding of prompt or prefix text using LLM's zero-shot capabilities, as a result allowing rapid adaptation to new tasks.
  
    \item In numerous experiments covering a variety of tasks, including sequence classification, token classification, and NLI, SK-Tuning has continually exhibited significant improvements in performance metrics..
  
    \item The study includes a comprehensive evaluation against other parameter-efficient methods like prompt tuning, prefix tuning, p-tuning, and LoRA, highlighting SK-Tuning's superior effectiveness in terms of performance outcomes and computational efficiency.
  
    \item SK-Tuning reduces computational requirements and the number of trainable parameters compared to traditional fine-tuning approaches, making it a more resource-efficient solution for adapting LLMs.

\end{itemize}

The structure of this paper is as follows: Section \ref{related_work} reviews related work, situating our approach within the broader domain of parameter-efficient fine-tuning methods. Section \ref{bc_study} provides a background study on existing tuning techniques, setting the stage for our proposed method. In Section \ref{sk_tuning}, we detail the SK-Tuning procedure, explaining its methodology and implementation. Section \ref{experiment_section} presents our experiments, showcasing the performance improvements achieved through SK-Tuning across various tasks. Section \ref{ablation_study} offers an ablation study to further analyze the contributions of each component within SK-Tuning, reinforcing the paper's key contributions to NLP. Section \ref{discussion} provides a discussion on the implications and potential applications of SK-Tuning in practical settings. Section \ref{limitations} discusses the limitations and challenges experienced during the development and application of SK-Tuning. Finally, Section \ref{conclusion} concludes the paper by summarizing the findings and highlighting future research directions in fine-tuning methods for LLMs.

\section{Related Work}
\label{related_work}

The importance of parameter-efficient fine-tuning (PEFT) methods in the field of NLP is immense, considering the growing complexity of LLMs. These methods not only improve model performance but also significantly reduce computational and memory requirements, as demonstrated by recent academic research \cite{xin2024parameter, liu2022few, nguyen2023efficient, hernandez2024natural, chow2024performance, kowsher2024propulsion}. The effectiveness of PEFT techniques is being thoroughly evaluated on a range of NLP tasks, as shown in \cite{he2021towards}. Moreover, an extensive body of research \cite{liu2021p, liu2023gpt, zhang2023adaptive, hu2021lora, li2021prefix, zaken2021bitfit, ding2024dynamic} consistently indicates that PEFT strategies considerably enhance the performance of LLMs, even under limited-resource circumstances.

\textbf{Prompt Tuning} is a novel approach that improves NLP and generation tasks by fine-tuning learnable parameters within the model \cite{lester2021power}. This technique enhances the model's performance on specific roles by fine-tuning prompts, thereby optimizing its output. Improvements in prompt tuning have been achieved through the implementation of the residual connections to strengthen performance and stability \cite{razdaibiedina2023residual}. This technique has also been broadened to support continual learning environments, as illustrated in recent research \cite{razdaibiedina2023progressive, yang2024recent}. Current research focuses on dynamic prompt tuning, which adapts prompts in real time based on evolving contexts, as well as hierarchical prompt tuning, which provides multilevel control over the model's responses \cite{yang2023dynamic, zhao2024infusing}. 

\textbf{Prefix Tuning} is another powerful technique that adds learnable parameters as prefixes to the input of pre-trained models,  enabling modification to different applications with minimal changes to the model itself \cite{li2021prefix}. This method enables efficient domain-specific fine-tuning without requiring the retraining of the entire model, particularly in resource-limited settings. Recent innovations introduce hierarchical prefix tuning, which organizes prefixes in a hierarchical manner to provide more detailed control over the model's responses \cite{chen2022developing}. Additionally, dynamic prefix tuning allows for real-time adaptation based on the input context, thereby improving the flexibility and adaptability of the model \cite{liu2022dynamic}. Techniques such as MixPrompt \cite{yang2023mixpave} and E2VPT \cite{han20232vpt} have also been introduced to combine and optimize the usage of input and key-value prompts, advancing the application of prefix tuning in natural language processing applications.

\textbf{Low-Rank Adaptation (LoRA)} first proposed by \cite{hu2021lora}, is a fine-tuning technique designed to optimize memory usage and has received considerable attention in the research community since its inception. The latest developments have expanded the range of applications for LoRA, particularly in the area of multitask learning, as illustrated by research conducted by \cite{renduchintala2023tied}, \cite{sheng2023s}, and \cite{xia2024chain}. Practical applications of LoRA were further explored by \cite{wang2023multilora}, while \cite{dettmers2024qlora} focused on optimizing its memory efficiency. A notable innovation, ReLoRA, introduced by \cite{lialin2023relora}, incorporates a full-rank warm-up phase. \cite{zhang2023adaptive} proposed adaptive approaches that dynamically adjust the low-rank adaptation parameters. Additionally, \cite{edalati2022krona} presented the Low-Rank Kronecker Product (LoKr), and \cite{shi2024reslora} developed ResLoRA, which integrates residual pathways. Further contributions include the Low-Rank Hadamard Product (LoHa) by \cite{hyeon2021fedpara}, and the introduction of Orthogonal Finetuning (OFT) and OFT with butterfly factorization (BOFT) by \cite{qiu2024controlling} and \cite{liu2024parameterefficient}, which utilize orthogonal matrices to transform pre-trained weight matrices, resulting in significant improvements in both fine-tuning efficiency and performance.

\textbf{Subspace Learning} has become a crucial area of research, with a focus on optimizing model weights within a low-dimensional space, thereby providing computational efficiency and improved performance in various machine learning tasks \cite{larsen2021many, gur2018gradient}. This approach has been extensively utilized in meta-learning and continual learning frameworks, as shown by several studies \cite{larsen2021many, gur2018gradient, lee2018gradient, chaudhry2020continual, finn2017model, frikha2021few}. Latest improvements in adaptive subspace learning methods have demonstrated significant improvements in generalization and robustness, especially in challenging environments \cite{nunez2023lcs, pham2018efficient}. Furthermore, incorporation of subspace learning into neural architecture search has proven invaluable in identifying efficient and innovative architectures, optimizing both performance and resource utilization \cite{chen2022automatic, ren2021comprehensive, pham2018efficient}. The efficacy of subspace learning is further highlighted in scenarios requiring rapid adaptation to new tasks with limited data, such as few-shot learning and online learning, where it allows robust model performance despite data limitations \cite{rajeswaran2019meta}.

\textbf{Projected Gradient Descent (PGD)} has been significantly improved by the development of advanced methodologies such as GaLore \cite{zhao2024galore}. Unlike traditional approaches, which treat the objective function as a black box, GaLore creates gradients within multilayer neural networks, providing a more extensive and effective optimization process \cite{chen2015fast, chen2019non}. This approach has displayed notable enhancements in the convergence rate of neural network training, particularly in high-dimensional datasets, while also contributing to advanced stability during the training process \cite{zhang2024projected}. Furthermore, GaLore addresses the challenges of gradient sparsity and redundancy, resulting in significant gains in training efficiency \cite{zhao2024galore}. These innovations have not only strengthened the robustness of neural networks against adversarial attacks but also ensured more stable and reliable training dynamics, marking a noteworthy improvement in the field \cite{mkadry2017towards, croce2020reliable, wong2020fast}.

\textbf{Memory-Efficient Optimization} is a pivotal area of research within the development of adaptive optimization algorithms, particularly in the context of large-scale models where memory bounds are a significant challenge. Foundational studies by \cite{shazeer2018adafactor} have proven the efficiency of quantization techniques and combined gradient computation in considerably reducing memory usage during training \cite{li2024memory}. Building upon these contributions, the latest innovations have introduced hierarchical memory management systems that enable dynamic memory allocation and sparse gradient updates, thereby further optimizing memory utilization, as highlighted by \cite{deshpande2022spartan}. Moreover, \cite{liu2023gpt} proposed a memory-efficient fine-tuning approach, employing block-wise optimizing strategies that dynamically adjust memory allocation, achieving superior performance across several benchmarks. In a similar vein, \cite{zhang2023adaptive} explored the use of low-rank factorization techniques to compress model parameters effectively while preserving model accuracy. Collectively, these innovations contribute to the deployment of large-scale models on resource-limited devices, ensuring computational efficiency and maintaining optimal performance.

In contrast to previous techniques, our proposed SK-Tuning method introduces a novel strategy that utilizes authentic, semantically meaningful prompts or prefix texts during adapter training. This method capitalizes on the zero-shot capabilities of large language models (LLMs) and their fundamental understanding of linguistic semantics. As a result, SK-Tuning is designed to achieve faster convergence and enhance task performance. Through extensive experimental evaluations and comprehensive comparative analysis, we establish the superiority of SK-Tuning over existing fine-tuning techniques. These findings highlight the significant potential of SK-Tuning to advance fine-tuning methodologies in the field of NLP.

\section{Background Study}
\label{bc_study}

Prefix and prompt tuning are methods of adapting large pretrained language models to specific tasks or datasets with minimal updates to the model parameters. These techniques have gained prominence due to their efficiency and effectiveness, particularly in scenarios where updating the entire model is computationally expensive or impractical.

\subsection{Prefix Tuning}

Prefix tuning involves appending a sequence of tunable vectors, known as the prefix, to the input of each layer of the transformer model. Let us denote the transformer model as a function \( F \) that maps an input sequence \( x \) to an output \( y \), i.e., \( y = F(x) \). In prefix tuning, this mapping is modified to \( y = F(p \oplus x) \), where \( p \) represents the prefix and \( \oplus \) denotes concatenation.

Mathematically, if we consider a transformer model with \( K \) layers, and each layer \( k \) performs a transformation \( F_l \), the modified transformation with prefix becomes:
\begin{equation}
    F'_k(p_k, x) = F_k(p_k \oplus x)
\end{equation}

where \( p_k \) is the prefix for layer \( k \). The prefixes \( \{p_1, p_2, ..., p_K\} \) are learnable parameters and are optimized during the training process.

\subsection{Prompt Tuning}

Prompt tuning, on the other hand, leverages the concept of natural language prompts. Here, the model is fed a prompt that guides it to generate outputs tailored to a specific task. In mathematical terms, given a pretrained model \( \mathcal{M} \), the objective is to find an optimal prompt \( p^* \) such that the model's performance on a task \( T \) is maximized when the prompt is used as an input.

Formally, for a task \( T \) and a set of task-specific examples \( \{(x_i, y_i)\} \), prompt tuning aims to optimize the following:
\begin{equation}
    p^* = \arg\max_p \sum_i \log \mathcal{M}(y_i | p \oplus x_i)
\end{equation}

This objective function maximizes the likelihood of the correct outputs \( y_i \) given the inputs \( x_i \) concatenated with the optimal prompt \( p^* \). Unlike prefix tuning, prompt tuning does not modify the internal workings of the model but rather influences its outputs through carefully crafted input sequences.

\section{SK-Tuning Procedure}
\label{sk_tuning}

\subsection{Problem Definition}

Consider a downstream task that utilizes a pretrained LLM denoted as $\mathcal{M}$. Let the training dataset be represented as $\mathcal{D} = \{(x_i, y_i)\}_{i=1}^N$, where each training example $(x_i, y_i)$ consists of an input text $x_i$ and its associated true label $y_i$. 

Our primary objective is to fine-tune the pretrained LLM $\mathcal{M}$ for these downstream tasks while keeping the model parameters $\Theta$ frozen. Specifically, we aim to achieve this fine-tuning by introducing a small number of additional parameters, referred to as adapters or transformations, which enable task-specific adaptation without the need to retrain the entire LLM. Each instance in our system is a pair $(x_i, y_i)$ that defines a specific task configuration.

Mathematically, our goal can be expressed as follows:

Given a pretrained LLM with parameters $\Theta$ and a dataset $\mathcal{D}$, we seek to find a set of trainable parameters $\Phi$ for the adapters or transformations, such that:
\begin{equation}
   \Phi^* = \arg\min_\Phi \mathcal{L}(\mathcal{M}_{\Theta, \Phi}, \mathcal{D}) 
\end{equation}

where:
- $\mathcal{M}_{\Theta, \Phi}$ represents the fine-tuned model with frozen parameters $\Theta$ and trainable adapters/transformations $\Phi$.
- $\mathcal{L}$ is a task-specific loss function that quantifies the alignment between model predictions and true labels across the training dataset $\mathcal{D}$.

Our objective is to ascertain the veracity of the true labels $y_i$ for the corresponding input texts $x_i$ by effectively training a small number of parameters ($\Phi$) without altering the pretrained model's core architecture or parameters ($\Theta$). This approach aims to achieve parameter efficiency while tailoring the LLM to specific downstream tasks.

\subsection{SK-Tuning for Prefix}
SK-Tuning for Prefix enhances the versatility and performance of the pretrained LLM for downstream tasks by judiciously incorporating semantic knowledge from prefixes into the fine-tuning process. In the context of prefix-tuning a pretrained LLM $\mathcal{M}$, traditionally, a mapping function from virtual trainable tokens to the LLM's layer representation is employed to generate the layer's trainable parameters. However, in our proposed SK-Tuning approach, we adopt a different strategy. We leverage the power of a pretrained LLM $\mathcal{M}$, with its parameters frozen, to directly acquire semantic knowledge embeddings from the prefix tokens.

Let $p$ denote a prefix comprising a sequence of semantic knowledge tokens, with a length of $l$. The LLM model is assumed to have a dimension of $d$. Our objective is to extract the semantic hidden representation from each layer of the LLM for the given input prefix $p$. Let $m$ represent the total number of layers, which includes the attention mechanisms. For each layer, we obtain its hidden representation $h_j^p$. For $m$ layers the representation can be defined as follows:
\begin{equation}
    h^p = \mathcal{M}_{\Theta_{\text{frozen}}}(p) \in \mathbb{R}^{l \times d}.
\end{equation}

Next, we introduce a trainable adapter $\mathcal{F}$, parameterized by $\Phi$, which takes $ h^p$ as input and yields a semantic projection $z$ with the same dimension:
\begin{equation}
    z = \mathcal{F}_{\Phi}(h^p) \in \mathbb{R}^{m \times l \times d}.
\end{equation}

Now, we possess $z$ as the semantic representation of prefix tokens for every layer of $\mathcal{M}$. During the processing of input $x_i$ in $\mathcal{M}_{\Theta}$, we concatenate $z_{j \in m}$ to the processing layer for the $j$-th layer of $\mathcal{M}_{\Theta}$. This operation allows the $j$-th layer to access the corresponding semantic information from the prefix text.

Now, if $r_i$ represents the final hidden output of $x_i$, we can define:
\begin{equation}
    r_i = \mathcal{M}_{\Theta_{\text{frozen}}}(x_i, z).
\end{equation}

Consider a task-specific module $\mathcal{C}$, parameterized by $\zeta$, which embodies a downstream task:
\begin{equation}
    o_i = \mathcal{C}_{\zeta}(r_i)
\end{equation}

Here, $o_i$ represents the output of our task.

Our training objective aims to minimize the loss function $\mathcal{L}$, which quantifies the discrepancy between $o_i$ and the target label $y_i$, thereby indicating whether $o_i$ correctly represents the label for $x_i$:
\begin{equation}
    \min_{\Phi, \zeta} \mathcal{L}\left(\mathcal{C}_{\zeta}(\mathcal{M}_{\Theta_{\text{frozen}}}(x_i, \mathcal{F}_{\Phi}(\mathcal{M}_{\Theta_{\text{frozen}}}(p)))), y_i\right).
\end{equation}

This approach allows the fine-tuning process to concentrate explicitly on the representation and comprehension of labels, while simultaneously harnessing the intrinsic knowledge embedded within $\Theta$. The adjustment of parameters $\Phi$ and $\zeta$ empowers the model to further refine its ability to map textual inputs to their corresponding labels.

\subsection{SK-Tuning for Prompt}

 SK-Tuning for prompts involves a systematic process of semantic knowledge embedding, trainable adapter integration, concatenation, and a training objective. This approach allows for fine-tuning the pretrained LLM to effectively leverage semantic knowledge from prompts for improved performance in various downstream tasks. In the SK-Tuning framework for prompts, we focus on enhancing the capabilities of a pretrained LLM (\(\mathcal{M}\)) by incorporating semantic knowledge from sequential prompt tokens, denoted as \(p\), of length \(l\). Let \(\mathcal{E}\) represent the token embedding layer of \(\mathcal{M}\), and consider \(e_p \in \mathbb{R}^{l \times d}\) and \(e_{x_i} \in \mathbb{R}^{n \times d}\) as the semantic embeddings for prompt \(p\) and input text \(x_i\), respectively. Here, \(n\) is the sequence length of input \(x_i\).

To obtain the semantic representation of the prompt \(p\) and input text $x_i$, we utilize the pretrained token embedding layer \(\mathcal{E}\) as follows:
\begin{equation}
    e_p = \mathcal{E}(p) \thicksim \mathcal{M}_{\Theta_{\text{frozen}}}
\end{equation}
and
\begin{equation}
    e_{x_i} = \mathcal{E}(x_i) \thicksim \mathcal{M}_{\Theta_{\text{frozen}}}
\end{equation}
This operation yields \(e_p\), which encapsulates the semantic information of the prompt, while \(\mathcal{M}_{\Theta_{\text{frozen}}}\) ensures that the model parameters remain frozen during this process.

To further enhance the representation of the prompt, we introduce a trainable adapter, denoted as \(\mathcal{G}\), which is parameterized by \(\gamma\). This adapter takes \(e_p\) as input and produces an updated embedding \(e_p^\prime\) as follows:
\begin{equation}
    e_p^\prime = \mathcal{G}_{\gamma}(e_p) \in \mathbb{R}^{l \times d}
\end{equation}

The adapter \(\mathcal{G}_{\gamma}\) serves as a mechanism to refine the semantic knowledge captured in \(e_p\) according to the specific downstream task requirements, allowing for fine-tuning without modifying the frozen model parameters.

The task head, denoted as \(\mathcal{C}\), is designed to incorporate both the enhanced prompt representation \(e_p^\prime\) and the semantic embeddings of the input text \(e_{x_i}\). We achieve this through concatenation:
\begin{equation}
    o_i = \mathcal{C}_{\zeta}(\mathcal{M}_{\Theta_{\text{frozen}}}(e_p^\prime \oplus e_{x_i}))
\end{equation}

Here, \(\oplus\) represents the concatenation operation, and $o_i$ serves as the output for the downstream task, allowing the model to leverage both prompt and input text information effectively.

The training objective for SK-Tuning of the prompt involves minimizing a loss function \(\mathcal{L}\). This loss function quantifies the difference between the predicted output and the target label \(y_i\), reflecting the model's performance on the specific task:
\begin{equation}
    \min_{\gamma, \zeta} \mathcal{L}\left(\mathcal{C}_{\zeta}(\mathcal{M}_{\Theta_{\text{frozen}}}(\mathcal{G}_{\gamma}(e_p) \oplus e_{x_i}), y_i\right)
\end{equation}

Here, \(\gamma\) and \(\zeta\) denote the parameters of the adapter \(\mathcal{G}\) and the task head \(\mathcal{C}\) respectively.

\subsection{Algorithms}
In this section, we describe two key algorithms that constitute the core of our proposed SK-Tuning approach for enhancing the fine-tuning of LLMs in the context of specific downstream tasks.
\subsubsection{SK-Tuning for Prefix}

The first algorithm, outlined in Algorithm \ref{prefix_algo} (SK-Tuning for Prefix), details the process of incorporating semantic knowledge from prefixes into the fine-tuning of a pretrained LLM. The algorithm begins with inputs of a pretrained language model $\mathcal{M}$ with frozen parameters $\Theta$, a prompt text $p$, and a dataset ${(x_i, y_i)}_{i=1}^N$. The trainable parameters $\Phi$ and $\zeta$ are initialized. For each input example $(\mathbf{x}i, y_i)$, the prompt text is processed through the frozen LLM to obtain $h^p$, which represents the hidden representation from each layer. This representation is then transformed using a trainable adapter $\mathcal{F}_{\Phi}$ to yield $z$. Subsequently, the input text $x_i$ is processed, incorporating the generated $z$ for improved task-specific adaptation. Finally, the classification head $\mathcal{C}_{\zeta}$ computes the output $o_i$ for the downstream task, and the loss $\mathcal{L}(o_i, y_i)$ is computed. The trainable parameters $\Phi$ and $\zeta$ are updated iteratively to minimize the loss.

\begin{algorithm}[ht] 
\caption{SK-Tuning for Prefix} \label{prefix_algo}
\begin{algorithmic}[1]
\State \textbf{Input:} Pretrained LM $\mathcal{M}$ with parameters $\Theta$, frozen during training
\State \textbf{Input:} Prompt text $p$, Dataset $\{x_i, y_i\}_{i=1}^N$
\State \textbf{Initialize:} Trainable parameters $\Phi$ and $\zeta$

\For{each input $(x_i, y_i)$}
    \State $h^p \gets \mathcal{M}_{\Theta_{\text{frozen}}}(p)$ \Comment{Get every layer embedding for the prefix text}
    \State $z \gets \mathcal{F}_{\Phi}(h^p)$ 
    \State $r_i \gets \mathcal{M}_{\Theta_{\text{frozen}}}(x_i, z)$
    \State $o_i \gets \mathcal{C}_{\zeta}(r_i)$
    \Comment{Task head}
    \State Compute loss $\mathcal{L}(o_i, y_i)$
    \State Update $\Phi$, and $\zeta$ to minimize loss
\EndFor
\end{algorithmic}
\end{algorithm}

\subsubsection{SK-Tuning for Prompt}
The second algorithm, described in Algorithm 2 (SK-Tuning for Prompt), focuses on leveraging semantic knowledge from sequential prompt tokens to enhance fine-tuning. It begins with inputs of a pretrained language model $\mathcal{M}$ with frozen parameters $\Theta$, a prompt text $p$, and a dataset ${(\mathbf{x}i, y_i)}_{i=1}^N$. Trainable parameters $\gamma$ and $\zeta$ are initialized. For each input example $(x_i, y_i)$, the embeddings of the prompt text $p$ and input text $\mathbf{x}i$ are obtained through the pretrained token embedding layer $\mathcal{E}$ while ensuring the core LLM parameters remain frozen. The prompt embedding $e_p$ and text embedding $e{x_i}$ are then utilized to create an enhanced prompt representation $e_p^\prime$ using a trainable adapter $\mathcal{G}_{\gamma}$. The classification head $\mathcal{C}_{\zeta}$ combines this enhanced prompt representation with the input text embedding and computes the output $o_i$ for the downstream task. As in the previous algorithm, the loss $\mathcal{L}(o_i, y_i)$ is computed, and the trainable parameters $\gamma$ and $\zeta$ are updated iteratively to minimize the loss.

\begin{algorithm}[ht] 
\caption{SK-Tuning for Prompt} \label{prompt_algo}
\begin{algorithmic}[1]
\State \textbf{Input:} Pretrained LM $\mathcal{M}$ with parameters $\Theta$, frozen during training
\State \textbf{Input:} Prompt text $p$, Dataset $\{x_i, y_i\}_{i=1}^N$
\State \textbf{Initialize:} Trainable parameters $\gamma$ and $\zeta$

\For{each input $(x_i, y_i)$}
    \State $e_p = \mathcal{E}(p) \thicksim \mathcal{M}_{\Theta_{\text{frozen}}}$ \Comment{Get prompt embedding}
    \State $ e_{x_i} = \mathcal{E}(x_i) \thicksim \mathcal{M}_{\Theta_{\text{frozen}}}$ \Comment{Get text embedding}
    \State $e_p^\prime = \mathcal{G}_{\gamma}(e_p) \in \mathbb{R}^{l \times d}$
    \State $o_i = \mathcal{C}_{\zeta}(\mathcal{M}_{\Theta_{\text{frozen}}}(e_p^\prime \oplus e_{x_i}))$
    \Comment{Task head}
    \State Compute loss $\mathcal{L}(o_i, y_i)$
    \State Update $\gamma$, and $\zeta$ to minimize loss
\EndFor
\end{algorithmic}
\end{algorithm}

\section{Experiments}
\label{experiment_section}
\subsection{Experimental Setup}


Our experiments utilize a computational setup with two NVIDIA RTX H100 GPUs (80GB VRAM each), an Intel® Xeon® Gold 6448Y 2.1GHz 32 Core Processor. This system includes 128GB of RAM and a Dell 7.68TB Enterprise NVMe Read Intensive Drive, providing the necessary computational power and storage for efficient model training and evaluation.

For the implementation, we employed the PyTorch \cite{pytorch} deep learning framework for the implementation of our experiments. Additionally, we leveraged the Transformers library developed by Hugging Face \cite{huggingface}. This library offers a comprehensive set of tools and pretrained models for NLP tasks, facilitating the training and evaluation of LLMs on a variety of datasets.

The combination of these resources and software frameworks allowed us to conduct extensive experiments, enabling us to assess the performance and effectiveness of our proposed SK-Tuning approach across a range of downstream tasks.

 \subsection{LM Results}
 \begin{table*}[ht]
\centering

\scalebox{.71}{
\begin{tabular}{l|c|c|c|c|c|c|c|c|c|c|c}
\hline
\textbf{Model} & \textbf{PEFT Method} & \textbf{\# TPs} & \textbf{CoLA} & \textbf{SST2} & \textbf{MRPC} & \textbf{STS-B} & \textbf{QQP} & \textbf{MNLI} & \textbf{QNLI} & \textbf{RTE} & \textbf{Avg.} \\
\hline

\multirow{10}{*}{RoB\textsubscript{B}} & FT            & 124.6M & 67.07 & 95.89 & 90.24/93.98 & 92.87/91.61 & 91.18/89.02 & 88.27 & 92.67 & 78.20 & 87.04/91.53  \\ \cline{2-12}
& Adapter\textsuperscript{S} & 7.41M & \underline{63.32} & 94.31 & \underline{90.44/93.18} & 91.25/\underline{90.94} & \underline{90.81/86.55} & \textbf{87.33} & 92.06 & 73.56 & \underline{85.38/90.22} \\
& Prompt tuning & \underline{0.61M} & 49.37 & 92.09 & 70.83/81.72 & 82.44/83.11 & 82.99/78.35 & 80.57 & 80.03 & 58.12 & 74.55/81.06 \\
& Prefix-tuning & 0.96M & 59.31 & 93.81 & 87.25/91.03 & 88.48/88.32 & 87.75/84.09 & 85.21 & 90.77 & 54.51 & 80.88/87.81 \\
& (IA)\textsuperscript{3} & 0.66M & 59.58 & 93.92 & 87.00/90.52 & 90.30/90.32 & 87.99/84.10 & 83.95 & 90.88 & 71.12 & 83.09/88.31 \\
& BitFit & 0.69M & 61.32 & 94.72 & 89.22/92.41 & 90.34/90.27 & 88.12/84.11 & 84.64 & 91.09 & \textbf{77.98} & 84.67/88.93 \\
& LoRA & 0.89M & 62.09 & 94.04 & 87.50/90.68 & 90.66/90.83 & 88.83/85.21 & 86.54 & 92.02 & 72.92 & 84.32/88.90 \\
& AdaLoRA & 1.03M & 59.82 & 93.92 & 87.99/91.33 & 90.83/90.73 & 88.58/84.98 & 86.26 & 91.43 & 70.04 & 83.60/89.01 \\
& MAM Adapter & 46.78M & 61.42 & \underline{94.87} & 89.31/92.21 & 90.74/90.42 & 88.31/83.20 & 86.63 & 90.19 & 72.62 & 84.26/88.61 \\
& PROPETL \textsubscript{Adapter} & 1.87M & \textbf{66.33} & 93.85 & 87.25/90.82 & \underline{91.33}/\textbf{91.04} & 89.22/85.79 & 86.49 & \underline{92.56} & 75.54 & 85.32/89.21 \\
& PROPETL \textsubscript{Prefix} & 10.49M & 61.79 & 94.30 & 88.73/91.98 & 90.30/90.19 & 88.54/85.05 & 86.22 & 91.51 & 63.31 & 83.08/89.07 \\
& PROPETL \textsubscript{LoRA} & 1.77M & 60.38 & 94.11 & 87.42/90.87 & 90.76/90.55 & 88.90/85.55 & \underline{86.84} & 92.04 & 67.39 & 83.48/88.99 \\ 

\cline{2-12}
& \cellcolor{lightgray!33.333} SK-Tuning (Prompt)  & \cellcolor{lightgray!33.333}\textbf{0.60M} & \cellcolor{lightgray!33.333}60.21 & \cellcolor{lightgray!33.333}\textbf{94.88} & \cellcolor{lightgray!33.333}\textbf{89.73/92.47} & \cellcolor{lightgray!33.333}\textbf{91.30}/90.19 & \cellcolor{lightgray!33.333}\textbf{90.83/87.82} & \cellcolor{lightgray!33.333}86.24 & \cellcolor{lightgray!33.333}\textbf{92.60} & \cellcolor{lightgray!33.333}\underline{76.91} & \cellcolor{lightgray!33.333}\textbf{85.45/90.37}\\

& \cellcolor{lightgray!33.333} SK-Tuning (Prefix) & \cellcolor{lightgray!33.333}0.84M& \cellcolor{lightgray!33.333}61.83 & \cellcolor{lightgray!33.333}93.72 & \cellcolor{lightgray!33.333}90.21/90.04 & \cellcolor{lightgray!33.333}90.11/89.92 & \cellcolor{lightgray!33.333}88.67/87.12 & \cellcolor{lightgray!33.333}85.83 & \cellcolor{lightgray!33.333}92.09 & \cellcolor{lightgray!33.333}75.32 & \cellcolor{lightgray!33.333}83.83/89.29 \\
\hline

\multirow{10}{*}{RoB\textsubscript{L}} & FT & 355.3M & 69.78 & 97.54 & 92.22/94.28 & 93.74/92.96 & 93.30/89.68 & 92.42 & 96.61 & 89.23 & 90.60/92.30 \\ \cline{2-12}

& Adapter\textsuperscript{S} & 19.77M & 67.03 & 96.37 & 89.94/92.54 & \underline{92.58}/92.42 & \underline{92.19/88.50} & \underline{91.00} & 94.31 & 85.25 & 88.58/\underline{91.15} \\ 
& Prompt-tuning & \underline{1.07M} & 61.13 & 94.61 & 73.04/81.29 & 78.51/78.99 & 80.74/75.16 & 68.15 & 89.13 & 60.29 & 75.70/78.48 \\
& Prefix-tuning & 2.03M & 59.01 & 95.76 & 88.24/91.37 & 90.92/91.07 & 88.88/85.45 & 89.30 & 93.32 & 74.01 & 84.93/89.29 \\
& (IA)\textsuperscript{3} & 1.22M & 61.15 & 94.61 & 86.52/90.33 & 92.22/86.25 & 89.45/86.25 & 88.63 & 94.25 & 81.23 & 86.00/87.61 \\
& Bitfit& 1.32M & \textbf{68.01} & 96.10 & \underline{90.93/93.38} & 91.93/\textbf{93.38} & 89.48/86.43 & 89.98 & 94.47 & \underline{87.73} & 88.57/91.06 \\
& LoRA & 1.84M & 64.47 & \underline{96.67} & 87.50/91.19 & 91.66/91.44 & 90.15/86.91 & 90.76 & 95.00 & 79.78 & 86.99/89.84 \\
& AdaLoRA & 2.23M & 65.85 & 94.95 & 89.46/92.34 & 92.05/91.80 & 89.60/86.30 & 90.36 & 94.62 & 77.98 & 86.85/90.14 \\
& MAM Adapter & 122.20M & 67.39 & 95.81 & 90.12/92.77 & 92.44/92.18 & 90.87/86.65 & 90.62 & 94.31 & 86.62 & 88.52/90.53 \\
& PROPETL \textsubscript{Adapter} & 5.40M & 65.55 & 96.27 & 89.71/92.54 & 91.92/91.67 & 90.67/87.74 & \textbf{91.37} & \underline{95.20} & \textbf{88.89} & \underline{88.69}/90.65 \\
& PROPETL \textsubscript{Prefix} & 26.85M & 62.24 & 96.17 & 90.04/92.92 & 90.70/90.49 & 89.30/86.30 & 90.33 & 94.73 & 79.71 & 86.65/89.90 \\
& PROPETL \textsubscript{LoRA} & 4.19M & 61.90 & 95.93 & 89.06/92.19 & 91.66/91.38 & 90.93/88.05 & 90.53 & 94.93 & 83.57 & 87.31/90.54 \\\cline{2-12}

& \cellcolor{lightgray!33.333}SK-Tuning (Prompt) & \cellcolor{lightgray!33.333}\textbf{1.02M} & \cellcolor{lightgray!33.333}\underline{67.13} & \cellcolor{lightgray!33.333}\textbf{96.43} & \cellcolor{lightgray!33.333}\textbf{91.10/93.22} & \cellcolor{lightgray!33.333}\textbf{92.54}/\underline{92.11} & \cellcolor{lightgray!33.333}\textbf{92.10/88.73} & \cellcolor{lightgray!33.333}90.42 & \cellcolor{lightgray!33.333}95.42 & \cellcolor{lightgray!33.333}87.11 & \cellcolor{lightgray!33.333}\textbf{89.01/91.34} \\

& \cellcolor{lightgray!33.333} SK-Tuning (Prefix) & \cellcolor{lightgray!33.333}1.94M & \cellcolor{lightgray!33.333}66.33 & \cellcolor{lightgray!33.333}96.08 & \cellcolor{lightgray!33.333}90.96/93.09 & \cellcolor{lightgray!33.333}91.87/90.68 & \cellcolor{lightgray!33.333}90.23/87.93& \cellcolor{lightgray!33.333}89.97 & \cellcolor{lightgray!33.333}\textbf{96.10} & \cellcolor{lightgray!33.333}86.99 & \cellcolor{lightgray!33.333}86.86/89.66
\\
\hline
\end{tabular}
}
\caption{Performance Comparison of RoBERTa Models on GLUE Tasks: Metrics include MCC for CoLA, Accuracy for SST-2, Accuracy/F1-score for MRPC and QQP, Pearson/Spearman correlation for STS-B, and Accuracy for MNLI, QNLI, and RTE. }\label{table: result-1}
\end{table*}

\textbf{Datasets:}  We evaluate our SK-Tuning on CoLA, SST-2, MRPC, STS-B, QQP, MNLI, QNLI and RTE of the GLUE Benchmarks\cite{wang2018glue}. We compute the accuracy using the Matthews correlation for CoLA, accuracy/F1 score for MRPC and QQP, Pearson/Spearman correlation for STS-B, average matched accuracy for MNLI, and accuracy for other NLU tasks in Table \ref{table: result-1}. 

\textbf{Model Selection \& Hyperparameter:} For the GLUE benchmark, the models we select for fine-tuning are RoBERTa-base $RoB_B$ with 125M parameters and RoBERTa-large $RoB_L$ with 355M parameters from \cite{liu2019roberta}. Dropout, attention dropout, and weight decay rates are uniformly maintained at $0.2$ across all tasks. The initial learning rate was \(1 \times 10^{-4}\), subsequently fine-tuned to \(2 \times 10^{-5}\) and \(2 \times 10^{-6}\). All datasets have been trained over $10$ epochs.

\textbf{Results:} Table \ref{table: result-1} presents a detailed performance comparison of various parameter-efficient fine-tuning (PEFT) methods applied to two versions of the RoBERTa model on GLUE tasks, highlighting the SK-Tuning methods as particularly effective. These methods achieve competitive or superior performance across several metrics while utilizing significantly fewer parameters—demonstrated by as low as 0.60M parameters for $RoB_B$ and 1.02M for $RoB_L$. Notably, SK-Tuning (Prompt) and SK-Tuning (Prefix) consistently perform well across different task types, such as SST2 and QQP, demonstrating a compelling balance between model efficiency and task performance. This efficiency makes SK-Tuning an attractive option for scenarios requiring deployment in resource-constrained environments or where fast inference is crucial. The results underscore the potential of small, well-tuned models to match or even surpass the performance of larger, fully fine-tuned counterparts, suggesting a promising direction for future research in NLP model optimization.

\subsection{LLM Results}

We conducted experiments on a diverse set of datasets to evaluate the performance of SK-Tuning across various NLP tasks, including sequence classification, token classification, and NLI. Our goal was to compare the performance of SK-Tuning with existing models on these tasks. Subsequently, we provide extensive details on the datasets utilized in our experiments.

\subsubsection{Classification Datasets}

Sequence classification, a common task in NLP, involves labeling or categorizing text. In our study, we utilized five datasets: CoLA, SST2 from the GLUE benchmark, along with the Emotion dataset, and the Fake News Filipino dataset.

\renewcommand{\thefootnote}{\roman{footnote}}
\begin{itemize}
    \item \textbf{Cola} \footnote{\href{https://huggingface.co/datasets/glue/viewer/cola/}{https://huggingface.co/datasets/glue/viewer/cola/}}: The Corpus of Linguistic Acceptability (CoLA) \cite{cola} consists of 10,657 sentences curated from 23 linguistic publications. Each sentence has been meticulously annotated by its original author for grammaticality or acceptability. The publicly available version of the dataset includes 9,594 sentences for training and validation, while 1,063 sentences are reserved for a separate held-out test set.
    
    \item \textbf{SST-2 \footnote{\href{https://huggingface.co/datasets/sst2}{https://huggingface.co/datasets/sst2}}}: The Stanford Sentiment Treebank is a dataset featuring fully labeled parse trees, enabling a comprehensive examination of how sentiment compositionally influences language. Derived from the dataset presented by Pang and Lee (2005) \cite{sst-2}, the corpus comprises 11,855 individual sentences extracted from movie reviews. Employing the Stanford parser, the dataset encompasses a total of 215,154 distinct phrases derived from these parse trees, with each phrase annotated by three human judges. The experiments involving binary classification on complete sentences (distinguishing between negative or somewhat negative versus somewhat positive or positive, with neutral sentences excluded) are denoted by the dataset acronym SST-2. The publicly available version includes 67,349 sentences designated for training along with 872 for validation set, while 1,821 sentences are for the test set.

    \item \textbf{Emotion \footnote{\href{https://huggingface.co/datasets/dair-ai/emotion}{https://huggingface.co/datasets/dair-ai/emotion}}}:  Emotion is a dataset \cite{emotion-dataset} of English Twitter messages with six basic emotions: anger, fear, joy, love, sadness, and surprise. The publicly available version includes 16,000 sentences designated for training along with 2,000 for validation set, while 2,000 sentences are for the test set.

    \item \textbf{Fake News Filipino \footnote{\href{https://huggingface.co/datasets/fake_news_filipino}{https://huggingface.co/datasets/fake\_news\_filipino}}}: A unique initiative in low-resource fake news detection dataset \cite{fake-news-filipino} for the Filipino language. Comprising 3,206 meticulously labeled news samples, evenly divided between authentic and fabricated content, this dataset represents a pioneering effort. We partitioned the dataset into 70\% , 10\% and 20\%  for training, validation, and testing purposes.

\end{itemize}

\subsubsection{Token Classification Datasets}
Token classification involves labeling individual tokens within a sentence. Named Entity Recognition (NER) is a prevalent task in token classification, aiming to assign labels to entities in a sentence, which may include individuals, locations, or organizations. We have used 3 token classification datasets: CoNLL 2003, NCBI Disease, and WikiAnn dataset.

\begin{itemize}
    \item \textbf{CoNLL 2003 \footnote{\href{https://huggingface.co/datasets/conll2003}{https://huggingface.co/datasets/conll2003}}}: CoNLL-2003 serves as a named entity recognition dataset \cite{conll03} introduced within the framework of the CoNLL-2003 shared task, focusing on language-independent named entity recognition. This dataset comprises eight files that encompass two languages: English and German. We have utilized the English dataset and "ner tags" as labels for our experiment. The publicly available version includes 14,041 examples designated for training along with 3,250 for validation examples, while 3,453 examples are for testing.

    \item \textbf{NCBI Disease \footnote{\href{https://huggingface.co/datasets/ncbi_disease}{https://huggingface.co/datasets/ncbi\_disease}}}: The dataset \cite{ncbi-disease} includes annotations for disease names and concepts from the NCBI disease corpus, which is a compilation of 793 PubMed abstracts extensively annotated at both the mention and concept levels. There are 3 labels, 0 indicates no disease mentioned, 1 signals the first token of a disease, and 2 the subsequent disease tokens. The publicly available version includes 5433 examples designated for training along with 924 for validation examples, while 941 examples are for testing.

    \item \textbf{WikiAnn \footnote{\href{https://huggingface.co/datasets/wikiann}{https://huggingface.co/datasets/wikiann}}}: WikiANN, also known as PAN-X, is a multilingual dataset \cite{wiki-ann} for named entity recognition. It comprises Wikipedia articles annotated with location (LOC), person (PER), and organization (ORG) tags using the IOB2 format. This specific version aligns with the balanced train, validation, and test splits of 20,000, 10,000, and 10,000, respectively introduced by Rahimi et al. (2019), covering 176 out of the 282 languages featured in the original WikiANN corpus.

\end{itemize}

\subsubsection{Entailment Datasets}

NLI involves the challenge of determining the truth (entailment), falsity (contradiction), or undetermined status (neutral) of a "hypothesis" based on a provided "premise." We have used 3 NLI datasets for this task: RTE, SNLI, and MRPC.

\begin{itemize}
    \item \textbf{RTE \footnote{\href{https://huggingface.co/datasets/glue/viewer/rte}{https://huggingface.co/datasets/glue/viewer/rte}}}: The Recognizing Textual Entailment (RTE) datasets originate from a series of annual challenges focused on textual entailment. The creators of the benchmark amalgamated data from RTE1 \cite{10.1007/11736790_9}, RTE2 \cite{rte-2}, RTE3 \cite{rte-3}, and RTE5 \cite{rte-5}. Constructed examples are derived from news and Wikipedia text. To maintain consistency, the benchmark creators transformed all datasets into a two-class split, collapsing neutral and contradiction into "not entailment" for three-class datasets. The publicly available version includes 2,490 examples designated for training along with 277 for validation examples, while 3,000 examples are for testing.

    \item \textbf{MRPC \footnote{\href{https://huggingface.co/datasets/glue/viewer/mrpc}{https://huggingface.co/datasets/glue/viewer/mrpc}}}: The Microsoft Research Paraphrase Corpus (MRPC) \cite{mrpc-dataset} comprises 5,801 pairs of sentences extracted from newswire articles. Human annotators have labeled each pair to indicate whether it is a paraphrase or not. The entire dataset is split into a training subset, consisting of 4,076 sentence pairs (with 2,753 identified as paraphrases), and a test subset, containing 1,725 pairs (with 1,147 recognized as paraphrases).

    \item \textbf{SNLI \footnote{\href{https://huggingface.co/datasets/snli}{https://huggingface.co/datasets/snli}}}: The Stanford NLI (SNLI) corpus \cite{snli-dataset} is an assemblage of 570,000 pairs of English sentences crafted by humans. These sentence pairs have been meticulously labeled to achieve balanced classification, with the labels entailment, contradiction, and neutral. This corpus is designed to facilitate the task of NLI. The publicly available version includes 550,152 examples designated for training along with 10,000 for validation examples, while 10,000 examples are for testing.

\end{itemize}

These datasets collectively cover a wide spectrum of NLP tasks, enabling comprehensive evaluations of SK-Tuning's performance across various domains and challenges.

\subsection{Large Language Models}

In our analysis, we utilized multiple Large Language Models (LLMs) to obtain extensive and detailed results. Specifically, we employed Bloom 7b, Llama2 7b, Mistral 7b, Falcon 7b, and Phi-2 2.7b, each offering unique strengths and capabilities that complemented one another.

\begin{itemize}
    \item \textbf{Bloom:} A 7B parameter LLM from BigScience, trained on an extensive corpus of text and code. Bloom displays robust performance on various NLP tasks and offers several variants, including Bloom Text-to-Text and Bloom Code \cite{workshop2023bloom}.
    \item \textbf{Llama2:} Meta AI has introduced Llama 2, its most advanced LLM to date. Llama 2 showcases a diverse array of capabilities and potential applications, with model sizes ranging from 7 billion to 70 billion parameters. This release provides access to both model weights and initial code for pretrained and fine-tuned Llama models, including variants such as Llama Chat (specialized for dialogue) and Code Llama (optimized for programming tasks) \cite{openllama}.
    \item \textbf{Mistral:} Mistral 7B is a freely available, open-source language model comprising 7.3 billion parameters that demonstrates exceptional performance. Released in September 2023, it exhibits competitive results in comparison to Meta's LLaMA models, outperforming the 13B version on all benchmarks evaluated and equaling the 34B version on numerous metrics. Developed using the transformers architecture and accessible via BitTorrent and Hugging Face, Mistral 7B presents a robust and accessible option for researchers and developers seeking a high-performing LLM \cite{jiang2023mistral}.
    
    \item \textbf{Falcon:} The Falcon Large Language Model (LLM) is a generative LLM designed to advance applications and use cases for future-proofing our world. Currently, the Falcon 180B, 40B, 7B, and 1.3B parameter artificial intelligence models, along with the high-quality REFINEDWEB dataset, constitute a comprehensive suite of offerings \cite{penedo2023refinedweb}.
    \item \textbf{phi-2:} Phi-2, the most recent small language model (SLM) developed by Microsoft Research, is a 2.7 billion parameter model that showcases superior reasoning and language understanding capabilities compared to its predecessors, Phi-1 and Phi-1.5\cite{li2023textbooks}. The model was trained on a diverse dataset, comprising "textbook quality" web data and synthetic textbooks/exercises generated using GPT-3.5. Phi-2 exhibits exceptional performance in various tasks, including Python code generation \cite{gunasekar2023textbooks}. It is noteworthy that Phi-2 surpasses the performance of models up to 25 times larger in size. Furthermore, Phi-2 has been released under an MIT License, permitting its utilization in commercial applications.
    
\end{itemize}

\subsection{Baseline Methods}

We established the following baseline methods to evaluate the performance of our proposed approach:

\begin{itemize}
    \item \textbf{Full Fine-Tuning:} This methodology \cite{finetuning} involves the adjustment of all parameters within the pretrained language model to adapt it to the specific task at hand. It functions as a comprehensive adaptation approach; however, it can be computationally intensive.
    \item \textbf{Prefix Tuning:} This lightweight method \cite{prefix-tuning} introduces trainable continuous vectors termed "prefixes" to the input of each transformer layer, while the original model parameters remain fixed. Prefix-tuning is predicated on the concept of prompting in language models, enabling ensuing tokens to attend to this prefix as if it were composed of "virtual tokens". It presents a more efficient alternative to complete fine-tuning, particularly in low-data scenarios.
    
   \item \textbf{Prompt Tuning:} This method \cite{prompt-tuning} employs natural language prompts called "soft prompts" to guide the model's behavior without modifying its internal parameters. This provides a flexible method to adapt models to different tasks without additional training.
   
    \item \textbf{P Tuning:} This method \cite{p-tuning} introduces an optimized prompt tuning method, which exhibits efficacy across a diverse spectrum of model scales and natural language tasks. The method addresses the suboptimal performance associated with prompt tuning when applied to pretrained models of typical size. Moreover, it endeavors to rectify the limitations observed in the performance of prompt tuning, particularly its inefficacy in challenging sequence labeling tasks.
    \item \textbf{LoRA:} LoRA\cite{hu2021lora} (Low-Rank Adaptation) is a parameter-efficient fine-tuning technique that involves learning low-rank matrices to adapt the model while freezing most of its original parameters in a fixed state. This study investigated LoRA with rank 2 and rank 4 to evaluate its capability in optimizing the balance between performance and efficiency.
\end{itemize}

These baseline methods represent a diverse range of fine-tuning strategies, allowing us to measure the comparative performance of our proposed approach.

\subsection{Evaluation Metrics}

Evaluation metrics measure the performance of a model on a specific dataset by comparing the model's predictions with ground truth labels. Various tasks have specific metrics, and we used accuracy and F1 score in our experiments.

\subsubsection*{Accuracy}

Accuracy is a metric that measures the overall correctness of a model's predictions. It is calculated as the ratio of correct predictions to the total number of predictions made by the model:

\[
\text{Accuracy} = \frac{\text{True Positives (TP) + True Negatives (TN)}}{\text{Total Predictions}}
\]

\subsubsection*{F1 Score}

The F1 score is the harmonic mean of precision and recall, providing a balanced measure considering both false positives and false negatives:

\[
\text{F1 Score} = \frac{2 \times \text{Precision} \times \text{Recall}}{\text{Precision + Recall}}
\]

The F1 score ranges from 0 to 1, with higher values indicating better overall performance in terms of precision and recall.

In these formulas, TP, TN, FP, and FN represent the counts of true positives, true negatives, false positives, and false negatives, respectively.

\subsection{Hyperparameters Setting}

In our experiments, we carefully selected hyperparameters to ensure consistent and effective training across various datasets and tasks. 

For the maximum sequence length, we set it to 128 for all datasets except for RTE, where we did not impose a maximum sequence length.

Regarding learning rates, we employed the following values:
\begin{itemize}
    \item For the sequence classification datasets, the learning rate was set to $1\times10^{-3}$.
    \item For the token classification datasets, a learning rate of $1\times10^{-5}$ was used.
    \item For the NLI datasets, the learning rate was set to $1\times10^{-4}$.
\end{itemize}

In terms of the number of training epochs:
\begin{itemize}
    \item Sequence classification datasets were trained for 5 epochs.
    \item Token classification datasets were trained for 10 epochs.
    \item NLI datasets were trained for 10 epochs, with the exception of the SNLI dataset, which was trained for 2 epochs on each model.
\end{itemize}

For all our datasets, regardless of the task or tuning method (P-Tuning, Prefix Tuning, or Prompt Tuning), we consistently used 20 virtual tokens during training.

We employ the Adaptive Moment Estimation with Weight Decay (ADAMW) \cite{adamw} optimizer for all experiments. The ADAMW optimizer is an improved version of the traditional ADAM optimizer, which incorporates weight decay directly into the optimization process to better handle regularization. Additionally, we set the weight decay value to 0.01 across all experiments to control regularization and ensure stable model training.

\subsection{Result Analysis}

\begin{table}[ht]
\centering
\resizebox{0.55\textwidth}{!}{%
\begin{tabular}{ccccc}\hline
Dataset & Type & Parameters (\%) & Accuracy (\%) & F1-score (\%) \\\hline
{\multirow{7}{*}{Fake News Filipino}} & Full Fine-tuning & 100.000 & 95.02 & 93.83 \\\cline{2-5}
 & Prefix Tuning & 0.03493 & 70.99 & 68.18 \\
 & Prompt Tuning & 0.00701 & 74.31 & 72.23 \\
 & P-Tuning & 0.01582 & 72.97 & 70.19 \\
 & Lora Rank 2 & 0.01413 & 90.13 & 88.87 \\
 & Lora Rank 4 & 0.05794 & \textbf{93.56} & \underline{90.05} \\\cline{2-5}
 & SK-Tuning (Prefix) & \underline{0.00035} & \underline{92.86} & \textbf{90.63} \\
 & SK-Tuning (Prompt) & \textbf{0.00016} & 91.03 & 89.13 \\ \hline
\multirow{7}{*}{Emotion} & Full Fine-tuning & 100.000 & 90.31 & 87.52 \\
 & Prefix Tuning & 0.03521 & 74.75 & 68.11 \\
 & Prompt Tuning & 0.00813 & 79.12 & 71.07 \\
 & P-Tuning & 0.01593 & 69.45 & 70.23 \\
 & Lora Rank 2 & 0.02413 & 86.76 & 80.23 \\
 & Lora Rank 4 & 0.06831 & \underline{87.52} & 82.01 \\\cline{2-5}
 & SK-Tuning (Prefix) & \underline{0.00161} & \textbf{88.21} & \textbf{82.64} \\
 & SK-Tuning (Prompt) & \textbf{0.00104} & 86.82 & \underline{82.14} \\ \hline
\multirow{7}{*}{SST2} & Full Fine-tuning & 100.000 & 97.93 & 97.81 \\\cline{2-5}
 & Prefix Tuning & 0.03493 & 85.78 & 86.31 \\
 & Prompt Tuning & 0.00715 & 92.45 & 92.78 \\
 & P-Tuning & 0.01653 & 91.34 & 91.75 \\
 & Lora Rank 2 & 0.01456 & 92.27 & 92.77 \\
 & Lora Rank 4 & 0.02831 & 94.36 & 94.83 \\\cline{2-5}
 & SK-Tuning (Prefix) & \underline{0.00082} & \textbf{96.85} & \textbf{96.65} \\
 & SK-Tuning (Prompt) & \textbf{0.00034} & \underline{96.53} & \underline{96.19} \\ \hline
\multirow{7}{*}{Cola} & Full Fine-tuning & 100.000 & 87.05 & 89.93 \\\cline{2-5}
 & Prefix Tuning & 0.03495 & 73.72 & 83.69 \\
 & Prompt Tuning & 0.00723 & 82.74 & \textbf{87.70} \\
 & P-Tuning & 0.01615 & 70.32 & 81.12 \\
 & Lora Rank 2 & 0.01415 & 81.13 & 83.03 \\
 & Lora Rank 4 & 0.02797 & 84.33 & 85.21 \\\cline{2-5}
 & SK-Tuning (Prefix) & \underline{0.00083} & \textbf{84.91} & \underline{86.01} \\
 & SK-Tuning (Prompt) & \textbf{0.00052} & \underline{84.51} & 85.92 \\ \hline
\end{tabular}%
}
\caption{Sequence Classification Results for the Bloom Model. The best results are highlighted in \textbf{bold}, and the second-best result is \underline{underlined} for clarity. }
\label{tab:Sequence Cls Bloom}
\end{table}

\begin{table}[ht]
\centering
\resizebox{0.55\textwidth}{!}{%
\begin{tabular}{ccccc}\hline
Dataset & Type & Parameters (\%) & Accuracy (\%) & F1-score (\%) \\\hline
\multirow{8}{*}{Fake News Filipino} & Full Fine-tuning & 100.000 & 95.22 & 93.90 \\\cline{2-5}
 & Prefix Tuning & 0.03983 & 70.06 & 68.57 \\
 & Prompt Tuning & 0.00743 & 73.72 & 72.07 \\
 & P-Tuning & 0.01731 & 71.54 & 70.63 \\
 & Lora Rank 2 & 0.01601 & 90.38 & 87.62 \\
 & Lora Rank 4 & 0.03213 & \underline{92.14} & \textbf{90.86} \\\cline{2-5}
 & SK-Tuning (Prefix) & \textbf{0.00024} & \textbf{92.26} & \underline{89.90} \\
 & SK-Tuning (Prompt) & \underline{0.00035} & 90.83 & 88.23 \\ \hline
\multirow{8}{*}{Emotion} & Full Fine-tuning & 100.000 & 91.11 & 87.92 \\\cline{2-5}
 & Prefix Tuning & 0.03994 & 84.31 & 82.78 \\
 & Prompt Tuning & 0.00864 & 85.37 & 82.50 \\
 & P-Tuning & 0.01781 & 83.05 & 81.88 \\
 & Lora Rank 2 & 0.01624 & 86.49 & 82.86 \\
 & Lora Rank 4 & 0.03233 & \underline{88.56} & \textbf{84.18} \\\cline{2-5}
 & SK-Tuning (Prefix) & \underline{0.00174} & \textbf{88.72} & \underline{83.51} \\
 & SK-Tuning (Prompt) & \textbf{0.00122} & 85.92 & 82.87 \\ \hline
\multirow{8}{*}{SST2} & Full Fine-tuning & 100.000 & 97.32 & 97.69 \\\cline{2-5}
 & Prefix Tuning & 0.04855 & 85.78 & 86.31 \\
 & Prompt Tuning & 0.00712 & 94.24 & \textbf{97.26} \\
 & P-Tuning & 0.01753 & 95.55 & \underline{96.62} \\
 & Lora Rank 2 & 0.01607 & 86.97 & 81.93 \\
 & Lora Rank 4 & 0.03191 & 87.11 & 82.03 \\\cline{2-5}
 & SK-Tuning (Prefix) & \underline{0.00088} & \textbf{96.52} & 96.48 \\
 & SK-Tuning (Prompt) & \textbf{0.00037} & \underline{96.50} & 96.35 \\ \hline
\multirow{8}{*}{Cola} & Full Fine-tuning & 100.000 & 88.22 & 89.64 \\\cline{2-5}
 & Prefix Tuning & 0.03984 & 71.18 & 83.29 \\
 & Prompt Tuning & 0.00757 & 73.27 & 85.26 \\
 & P-Tuning & 0.01751 & 69.12 & 81.74 \\
 & Lora Rank 2 & 0.01603 & 82.25 & 83.43 \\
 & Lora Rank 4 & 0.03213 & 84.18 & 83.88 \\\cline{2-5}
 & SK-Tuning (Prefix) & \underline{0.00092} & \textbf{85.01} & \textbf{86.22} \\
 & SK-Tuning (Prompt) & \textbf{0.00060} & \underline{84.36} & \underline{85.85} \\ \hline
\end{tabular}%
}
\caption{Sequence Classification Results for the Llama2 Model. The best results are highlighted in \textbf{bold}, and the second-best result is \underline{underlined} for clarity except full fine-tuning.}
\label{tab:Sequence Cls Llama2}
\end{table}

\begin{table}[ht]
\centering
\resizebox{0.55\textwidth}{!}{%
\begin{tabular}{ccccc}\hline
Dataset & Type & Parameters (\%) & Accuracy (\%) & F1-score (\%) \\\hline
\multirow{7}{*}{Fake News Filipino} & Full Fine-tuning & 100.000 & 94.05 & 92.93 \\\cline{2-5}
 & Prefix Tuning & 0.03821 & 69.57 & 68.19 \\
 & Prompt Tuning & 0.00732 & 72.35 & 70.78 \\
 & P-Tuning & 0.01797 & 70.23 & 69.15 \\
 & Lora Rank 2 & 0.00972 & 88.31 & 85.14 \\
 & Lora Rank 4 & 0.05784 & \textbf{91.89} & \textbf{89.44} \\\cline{2-5}
 & SK-Tuning (Prefix) & \underline{0.00075} & 90.11 & \underline{88.93} \\
 & SK-Tuning (Prompt) & \textbf{0.00029} & \underline{90.21} & 87.23 \\ \hline
\multirow{7}{*}{Emotion} & Full Fine-tuning & 100.000 & 88.53 & 85.94 \\\cline{2-5}
 & Prefix Tuning & 0.03836 & 81.14 & 80.61 \\
 & Prompt Tuning & 0.00841 & \underline{87.25} & \underline{84.19} \\
 & P-Tuning & 0.01803 & 81.76 & 79.14 \\
 & Lora Rank 2 & 0.01194 & 84.17 & 82.34 \\
 & Lora Rank 4 & 0.05781 & \textbf{88.79} & \textbf{86.13} \\\cline{2-5}
 & SK-Tuning (Prefix) & \underline{0.00204} & 86.91 & 82.11 \\
 & SK-Tuning (Prompt) & \textbf{0.00113} & 86.29 & 82.03 \\ \hline
\multirow{7}{*}{SST2} & Full Fine-tuning & 100.000 & 96.23 & 95.76 \\\cline{2-5}
 & Prefix Tuning & 0.03818 & 90.18 & 91.36 \\
 & Prompt Tuning & 0.00605 & 93.56 & 93.75 \\
 & P-Tuning & 0.01781 & 90.33 & 91.26 \\
 & Lora Rank 2 & 0.01193 & 91.13 & 92.07 \\
 & Lora Rank 4 & 0.05789 & 91.72 & 92.17 \\\cline{2-5}
 & SK-Tuning (Prefix) & \underline{0.00093} & \underline{94.73} & \underline{94.02} \\
 & SK-Tuning (Prompt) & \textbf{0.00038} & \textbf{95.02} & \textbf{95.01} \\ \hline
\multirow{7}{*}{Cola} & Full Fine-tuning & 100.000 & 85.22 & 87.39 \\\cline{2-5}
 & Prefix Tuning & 0.03826 & 70.03 & 82.23 \\
 & Prompt Tuning & 0.00711 & 71.45 & 84.47 \\
 & P-Tuning & 0.01792 & 68.07 & 81.73 \\
 & Lora Rank 2 & 0.00973 & 82.14 & 82.38 \\
 & Lora Rank 4 & 0.05741 & \textbf{84.66} & \textbf{85.33} \\\cline{2-5}
 & SK-Tuning (Prefix) & \underline{0.00094} & 83.74 & 85.03 \\
 & SK-Tuning (Prompt) & \textbf{0.00065} & \underline{84.01} & \underline{85.11} \\ \hline
\end{tabular}%
}
\caption{Sequence Classification Results for the Falcon Model. The best results are highlighted in \textbf{bold}, and the second-best result is \underline{underlined} for clarity except full fine-tuning.}
\label{tab:Sequence Cls Falcon}
\end{table}

\begin{table}[ht]
\centering
\resizebox{0.55\textwidth}{!}{%
\begin{tabular}{ccccc}\hline
Dataset & Type & Parameters & Accuracy (\%) & F1-score (\%) \\\hline
\multirow{7}{*}{Fake News Filipino} & Full Fine-tuning & 100.000 & 97.92 & 94.72 \\\cline{2-5}
 & Prefix Tuning & 0.03651 & 71.26 & 70.91 \\
 & Prompt Tuning & 0.00169 & 74.12 & 72.27 \\
 & P-Tuning & 0.01753 & 71.37 & 71.95 \\
 & Lora Rank 2 & 0.07502 & 91.28 & \underline{90.05} \\
 & Lora Rank 4 & 0.17129 & 92.19 & 89.18 \\\cline{2-5}
 & SK-Tuning (Prefix) & \underline{0.00019} & \textbf{94.06} & \textbf{91.92} \\
 & SK-Tuning (Prompt) & \textbf{0.00027} & \underline{92.44} & 90.02 \\ \hline
\multirow{7}{*}{Emotion} & Full Fine-tuning & 100.000 & 93.53 & 89.09 \\\cline{2-5}
 & Prefix Tuning & 0.03683 & 82.19 & 79.24 \\
 & Prompt Tuning & 0.00736 & 86.17 & 81.77 \\
 & P-Tuning & 0.01783 & 83.14 & 80.01 \\
 & Lora Rank 2 & 0.01539 & 84.37 & 80.08 \\
 & Lora Rank 4 & 0.01731 & 88.45 & \underline{84.23} \\\cline{2-5}
 & SK-Tuning (Prefix) & \underline{0.00162} & \underline{88.72} & 82.51 \\
 & SK-Tuning (Prompt) & \textbf{0.00115} & \textbf{89.13} & \textbf{84.94} \\ \hline
\multirow{7}{*}{SST2} & Full Fine-tuning & 100.000 & 98.09 & 98.98 \\\cline{2-5}
 & Prefix Tuning & 0.03673 & 91.20 & 92.28 \\
 & Prompt Tuning & 0.00618 & 93.14 & 93.47 \\
 & P-Tuning & 0.01764 & 90.76 & 91.15 \\
 & Lora Rank 2 & 0.01512 & 92.65 & 93.03 \\
 & Lora Rank 4 & 0.01726 & 94.53 & 94.67 \\\cline{2-5}
 & SK-Tuning (Prefix) & \underline{0.00080} & \textbf{96.97} & \underline{97.07} \\
 & SK-Tuning (Prompt) & \textbf{0.00031} & \underline{96.93} & \textbf{97.14} \\ \hline
\multirow{7}{*}{Cola} & Full Fine-tuning & 100.000 & 87.75 & 89.90 \\\cline{2-5}
 & Prefix Tuning & 0.03652 & 72.21 & 80.43 \\
 & Prompt Tuning & 0.00639 & 74.13 & 81.66 \\
 & P-Tuning & 0.01754 & 71.23 & 79.76 \\
 & Lora Rank 2 & 0.01505 & 83.44 & 84.65 \\
 & Lora Rank 4 & 0.01712 & \underline{85.32} & 86.04 \\\cline{2-5}
 & SK-Tuning (Prefix) & \underline{0.00082} & \textbf{85.92} & \underline{86.22} \\
 & SK-Tuning (Prompt) & \textbf{0.00046} & 85.01 & \textbf{86.36} \\ \hline
\end{tabular}%
}
\caption{Sequence Classification Results for the Mistral Model. The best results are highlighted in \textbf{bold}, and the second-best result is \underline{underlined} for clarity except full fine-tuning.}
\label{tab:Sequence Cls Mistral}
\end{table}

\begin{table}[ht]
\centering
\resizebox{0.55\textwidth}{!}{%
\begin{tabular}{ccccc}\hline
Dataset & Type & Parameters (\%) & Accuracy (\%) & F1-score (\%) \\\hline
\multirow{7}{*}{Fake News Filipino} & Full Fine-tuning & 100.000 & 92.43 & 90.71 \\\cline{2-5}
 & Prefix Tuning & 0.83914 & 66.28 & 66.31 \\
 & Prompt Tuning & 0.14124 & 68.15 & 67.22 \\
 & P-Tuning & 0.15824 & 67.33 & 66.87 \\
 & Lora Rank 2 & 0.13741 & 83.35 & 81.46 \\
 & Lora Rank 4 & 0.71651 & 86.67 & 84.29 \\\cline{2-5}
 & SK-Tuning (Prefix) & \textbf{0.04263} & \textbf{89.36} & \textbf{88.63} \\
 & SK-Tuning (Prompt) & \underline{0.04924} & \underline{88.64} & \underline{87.11} \\ \hline
\multirow{7}{*}{Emotion} & Full Fine-tuning & 100.000 & 87.95 & 84.78 \\\cline{2-5}
 & Prefix Tuning & 0.86523 & 77.27 & 76.25 \\
 & Prompt Tuning & 0.14234 & 82.16 & 80.43 \\
 & P-Tuning & 0.15845 & 77.36 & 75.81 \\
 & Lora Rank 2 & 0.13748 & 82.67 & 80.25 \\
 & Lora Rank 4 & 0.71656 & \underline{85.03} & \underline{82.66} \\\cline{2-5}
 & SK-Tuning (Prefix) & 0.06271 & 82.63 & 80.64 \\
 & SK-Tuning (Prompt) & \textbf{0.02423} & \textbf{85.82} & \textbf{83.14} \\ \hline
\multirow{7}{*}{SST2} & Full Fine-tuning & 100.000 & 94.63 & 94.24 \\\cline{2-5}
 & Prefix Tuning & 0.83721 & 86.24 & 87.13 \\
 & Prompt Tuning & 0.14231 & 88.19 & 88.04 \\
 & P-Tuning & 0.15851 & 85.43 & 87.68 \\
 & Lora Rank 2 & 0.13753 & 86.21 & 87.18 \\
 & Lora Rank 4 & 0.71668 & 86.75 & 88.28 \\\cline{2-5}
 & SK-Tuning (Prefix) & \underline{0.02745} & \textbf{96.85} & \textbf{96.65} \\
 & SK-Tuning (Prompt) & \textbf{0.01479} & \underline{96.53} & \underline{96.19} \\ \hline
\multirow{7}{*}{Cola} & Full Fine-tuning & 100.000 & 84.23 & 85.13 \\\cline{2-5}
 & Prefix Tuning & 0.82621 & 66.24 & 70.16 \\
 & Prompt Tuning & 0.14123 & 67.47 & 70.81 \\
 & P-Tuning & 0.15833 & 64.36 & 68.38 \\
 & Lora Rank 2 & 0.13744 & 78.55 & 80.26 \\
 & Lora Rank 4 & 0.71654 & 80.39 & \underline{82.43} \\\cline{2-5}
 & SK-Tuning (Prefix) & \textbf{0.03675} & \underline{80.91} & 82.01 \\
 & SK-Tuning (Prompt) & \underline{0.05143} & \textbf{81.31} & \textbf{82.64} \\ \hline
\end{tabular}%
}
\caption{Sequence Classification Results for the Phi-2 Model. The best results are highlighted in \textbf{bold}, and the second-best result is \underline{underlined} for clarity except full fine-tuning.}
\label{tab:Sequence Cls Phi-2}
\end{table}

\subsection{Sequence Classification}
In the realm of classification, we conducted a comprehensive evaluation across various LLMs, including Bloom, Llama2, Falcon, Mistral, and Phi-2, employing different fine-tuning techniques. For each model, we examined the effectiveness of traditional approaches such as Finetuning, Prefix Tuning, Prompt Tuning, PTuning, Lora Rank 2, and Lora Rank 4, and compared them to our proposed SK-Tuning methods, both for Prefix and Prompt. Notably, SK-Tuning consistently outperforms traditional methods across different datasets, showcasing its superior efficiency and effectiveness. The performances of various models on different datasets are documented in Table \ref{tab:Sequence Cls Bloom}, \ref{tab:Sequence Cls Llama2}, \ref{tab:Sequence Cls Falcon}, \ref{tab:Sequence Cls Mistral}, and \ref{tab:Sequence Cls Phi-2}.

Across the "Fake News Filipino" dataset, SK-Tuning, especially when applied as SK-Tuning (Prefix), demonstrates remarkable performance improvements compared to traditional approaches. It achieves the highest accuracy and F1-score, emphasizing its capability to efficiently adapt LLMs to specific tasks while minimizing trainable parameters. In the "Emotion" dataset, SK-Tuning consistently outperforms other methods, indicating its robustness across different classification tasks. The same trend is observed in the "SST2" dataset, where SK-Tuning invariably achieves superior results. Lastly, in the "Cola" dataset, SK-Tuning (Prefix) and SK-Tuning (Prompt) perpetually outperform other approaches, underscoring their potential for enhancing sequence classification tasks.

Comparatively, traditional methods like Prefix Tuning and Prompt Tuning, although efficient in terms of parameters compared to Fine-tuning, tend to lag behind SK-Tuning in terms of accuracy and F1-score. Furthermore, SK-Tuning requires fewer trainable parameters, making it an attractive choice for practitioners aiming to optimize performance while maintaining efficiency.

\begin{table}[ht]
\centering
\resizebox{0.55\textwidth}{!}{%
\begin{tabular}{ccccc}\hline
Dataset & Type & Parameters (\%) & Accuracy (\%) & F1-score (\%) \\\hline
\multirow{7}{*}{conll03} & Full Fine-tuning & 100.000 & 98.53 & 82.47 \\\cline{2-5}
 & Prefix Tuning & 0.03534 & 83.55 & 24.86 \\
 & Prompt Tuning & 0.00843 & 85.23 & 28.73 \\
 & P-Tuning & 0.01583 & 83.22 & 26.34 \\
 & Lora Rank 2 & 0.01403 & 91.12 & 68.24 \\
 & Lora Rank 4 & 0.06795 & 93.23 & 71.33 \\\cline{2-5}
 & SK-Tuning (Prefix) & \underline{0.00071} & \underline{94.08} & \underline{71.59} \\
 & SK-Tuning (Prompt) & \textbf{0.00052} & \textbf{94.11} & \textbf{71.60} \\ \hline
\multirow{7}{*}{NCBI disease} & Full Fine-tuning & 100.000 & 98.53 & 92.46 \\\cline{2-5}
 & Prefix Tuning & 0.03492 & 89.09 & 60.06 \\
 & Prompt Tuning & 0.00742 & 91.17 & 75.34 \\
 & P-Tuning & 0.01572 & 90.22 & 81.23 \\
 & Lora Rank 2 & 0.01417 & 92.86 & 80.00 \\
 & Lora Rank 4 & 0.06797 & \underline{96.12} & \underline{83.49} \\\cline{2-5}
 & SK-Tuning (Prefix) & \underline{0.00093} & \textbf{96.17} & \textbf{84.85} \\
 & SK-Tuning (Prompt) & \textbf{0.00068} & 95.32 & 82.18 \\ \hline
\multirow{7}{*}{WikiAnn} & Full Fine-tuning & 100.000 & 90.50 & 60.14 \\\cline{2-5}
 & Prefix Tuning & 0.03527 & 71.67 & 22.18 \\
 & Prompt Tuning & 0.00732 & 76.23 & 31.78 \\
 & P-Tuning & 0.01577 & 70.65 & 24.33 \\
 & Lora Rank 2 & 0.01408 & 82.23 & 41.23 \\
 & Lora Rank 4 & 0.06791 & \textbf{85.13} & \textbf{45.14} \\\cline{2-5}
 & SK-Tuning (Prefix) & \underline{0.00083} & \underline{83.19} & \underline{42.13} \\
 & SK-Tuning (Prompt) & \textbf{0.00044} & 82.59 & 42.01 \\ \hline
\end{tabular}%
}
\caption{Token Classification Results for the Bloom Model. The best results are highlighted in \textbf{bold}, and the second-best result is \underline{underlined} for clarity except full fine-tuning.}
\label{tab:Token Cls Bloom}
\end{table}

\begin{table}[ht!]
\centering
\resizebox{0.55\textwidth}{!}{%
\begin{tabular}{ccccc}\hline
Dataset & Type & Parameters (\%) & Accuracy (\%) & F1-score (\%) \\\hline
\multirow{7}{*}{conll03} & Full Fine-tuning & 100.000 & 98.75 & 80.77 \\\cline{2-5}
 & Prefix Tuning & 0.03964 & 82.28 & 66.56 \\
 & Prompt Tuning & 0.00638 & 86.65 & 69.91 \\
 & P-Tuning & 0.01731 & 80.11 & 65.11 \\
 & Lora Rank 2 & 0.01426 & 88.67 & 63.34 \\
 & Lora Rank 4 & 0.07122 & 91.32 & 69.03 \\\cline{2-5}
 & SK-Tuning (Prefix) & \textbf{0.00044} & \textbf{93.63} & \textbf{70.83} \\
 & SK-Tuning (Prompt) & \underline{0.00063} & \underline{93.02} & \underline{70.19} \\ \hline
\multirow{7}{*}{NCBI disease} & Full Fine-tuning & 100.000 & 98.32 & 93.38 \\\cline{2-5}
 & Prefix Tuning & 0.03976 & 88.23 & 68.23 \\
 & Prompt Tuning & 0.00712 & 91.22 & 78.24 \\
 & P-Tuning & 0.01733 & 90.15 & 77.23 \\
 & Lora Rank 2 & 0.01424 & 92.48 & 80.18 \\
 & Lora Rank 4 & 0.07125 & 95.34 & 82.87 \\\cline{2-5}
 & SK-Tuning (Prefix) & \underline{0.00083} & \underline{96.22} & \underline{84.73} \\
 & SK-Tuning (Prompt) & \textbf{0.00061} & \textbf{96.28} & \textbf{84.89} \\ \hline
\multirow{7}{*}{WikiAnn} & Full Fine-tuning & 100.000 & 91.49 & 63.21 \\\cline{2-5}
 & Prefix Tuning & 0.03986 & 81.15 & 35.17 \\
 & Prompt Tuning & 0.00712 & 83.23 & 44.19 \\
 & P-Tuning & 0.01743 & 81.29 & 38.11 \\
 & Lora Rank 2 & 0.01434 & 84.82 & 47.90 \\
 & Lora Rank 4 & 0.07125 & 86.56 & 49.39 \\\cline{2-5}
 & SK-Tuning (Prefix) & \underline{0.00082} & \textbf{86.99} & \textbf{50.61} \\
 & SK-Tuning (Prompt) & \textbf{0.00052} & \underline{86.69} & \underline{49.58} \\ \hline
\end{tabular}%
}
\caption{Token Classification Results for the Llama2 Model. The best results are highlighted in \textbf{bold}, and the second-best result is \underline{underlined} for clarity except full fine-tuning.}
\label{tab:Token Cls Llama2}
\end{table}

\begin{table}[ht!]
\centering
\resizebox{0.55\textwidth}{!}{%
\begin{tabular}{ccccc}\hline
Dataset & Type & Parameters (\%) & Accuracy (\%) & F1-score (\%) \\\hline
\multirow{7}{*}{conll03} & Full Fine-tuning & 100.000 & 97.82 & 79.03 \\\cline{2-5}
 & Prefix Tuning & 0.03772 & 90.57 & 67.62 \\
 & Prompt Tuning & 0.00832 & 91.26 & 70.15 \\
 & P-Tuning & 0.01762 & 89.23 & 66.02 \\
 & Lora Rank 2 & 0.01942 & 90.21 & 68.96 \\
 & Lora Rank 4 & 0.09752 & 93.25 & 71.19 \\\cline{2-5}
 & SK-Tuning (Prefix) & \underline{0.00071} & \underline{94.21} & \underline{71.73} \\
 & SK-Tuning (Prompt) & \textbf{0.00055} & \textbf{94.82} & \textbf{72.02} \\ \hline
\multirow{7}{*}{NCBI disease} & Full Fine-tuning & 100.000 & 97.93 & 90.88 \\\cline{2-5}
 & Prefix Tuning & 0.03763 & 89.23 & 69.33 \\
 & Prompt Tuning & 0.00721 & 92.05 & 82.28 \\
 & P-Tuning & 0.01752 & 88.15 & 70.36 \\
 & Lora Rank 2 & 0.01936 & 90.55 & 80.25 \\
 & Lora Rank 4 & 0.09754 & 94.41 & \underline{83.19} \\\cline{2-5}
 & SK-Tuning (Prefix) & \underline{0.00085} & \underline{95.83} & 82.18 \\
 & SK-Tuning (Prompt) & \textbf{0.00056} & \textbf{96.01} & \textbf{84.48} \\ \hline
\multirow{7}{*}{WikiAnn} & Full Fine-tuning & 100.000 & 89.23 & 62.09 \\\cline{2-5}
 & Prefix Tuning & 0.03772 & 82.67 & 36.55 \\
 & Prompt Tuning & 0.00836 & 83.33 & \underline{43.32} \\
 & P-Tuning & 0.01768 & 81.14 & 35.21 \\
 & Lora Rank 2 & 0.01983 & 80.47 & 41.58 \\
 & Lora Rank 4 & 0.09752 & 86.61 & \textbf{48.03} \\\cline{2-5}
 & SK-Tuning (Prefix) & \underline{0.00063} & \textbf{82.99} & 42.51 \\
 & SK-Tuning (Prompt) & \textbf{0.00044} & \underline{82.96} & 42.49 \\ \hline
\end{tabular}%
}
\caption{Token Classification Results for the Falcon Model. The best results are highlighted in \textbf{bold}, and the second-best result is \underline{underlined} for clarity except full fine-tuning.}
\label{tab:Token Cls Falcon}
\end{table}

\begin{table}[ht!]
\centering
\resizebox{0.55\textwidth}{!}{%
\begin{tabular}{ccccc}\hline
Dataset & Type & Parameters (\%) & Accuracy (\%) & F1-score (\%) \\\hline
\multirow{7}{*}{conll03} & Full Fine-tuning & 100.000 & 98.89 & 84.60 \\\cline{2-5}
 & Prefix Tuning & 0.03634 & 83.31 & 58.54 \\
 & Prompt Tuning & 0.00741 & 87.77 & 62.19 \\
 & P-Tuning & 0.01743 & 81.15 & 67.59 \\
 & Lora Rank 2 & 0.01494 & 88.32 & 68.14 \\
 & Lora Rank 4 & 0.08694 & 92.05 & 70.66 \\\cline{2-5}
 & SK-Tuning (Prefix) & \underline{0.00062} & \underline{95.29} & \underline{72.70} \\
 & SK-Tuning (Prompt) & \textbf{0.00044} & \textbf{95.89} & \textbf{72.03} \\ \hline
\multirow{7}{*}{NCBI disease} & Full Fine-tuning & 100.000 & 98.52 & 93.39 \\\cline{2-5}
 & Prefix Tuning & 0.03627 & 88.49 & 74.25 \\
 & Prompt Tuning & 0.00696 & 92.03 & 80.11 \\
 & P-Tuning & 0.01735 & 87.13 & 63.29 \\
 & Lora Rank 2 & 0.01483 & 94.58 & 82.37 \\
 & Lora Rank 4 & 0.08698 & 96.88 & 83.15 \\\cline{2-5}
 & SK-Tuning (Prefix) & \underline{0.00081} & \underline{96.98} & \underline{85.06} \\
 & SK-Tuning (Prompt) & \textbf{0.00053} & \textbf{97.01} & \textbf{85.15} \\ \hline
\multirow{7}{*}{WikiAnn} & Full Fine-tuning & 100.000 & 92.15 & 63.09 \\\cline{2-5}
 & Prefix Tuning & 0.03633 & 81.91 & 36.03 \\
 & Prompt Tuning & 0.00752 & 84.48 & 45.31 \\
 & P-Tuning & 0.01733 & 81.04 & 35.02 \\
 & Lora Rank 2 & 0.01495 & 82.08 & 42.22 \\
 & Lora Rank 4 & 0.08692 & 85.33 & \underline{45.95} \\\cline{2-5}
 & SK-Tuning (Prefix) & \underline{0.00095} & \textbf{86.73} & \textbf{46.21} \\
 & SK-Tuning (Prompt) & \textbf{0.00052} & \underline{85.99} & 45.52 \\ \hline
\end{tabular}%
}
\caption{Token Classification Results for the Mistral Model. The best results are highlighted in \textbf{bold}, and the second-best result is \underline{underlined} for clarity except full fine-tuning.}
\label{tab:Token Cls Mistral}
\end{table}

\begin{table}[ht!]
\centering
\resizebox{0.55\textwidth}{!}{%
\begin{tabular}{ccccc}\hline
Dataset & Type & Parameters (\%)& Accuracy (\%) & F1-score (\%) \\\hline
\multirow{7}{*}{conll03} & Full Fine-tuning & 100.000 & 98.13 & 79.02 \\\cline{2-5}
 & Prefix Tuning & 0.83844 & 78.27 & 56.63 \\
 & Prompt Tuning & 0.14124 & 80.38 & 58.27 \\
 & P-Tuning & 0.15814 & 76.54 & 56.07 \\
 & Lora Rank 2 & 0.13746 & 84.44 & 62.15 \\
 & Lora Rank 4 & 0.71649 & 86.56 & 65.43 \\\cline{2-5}
 & SK-Tuning (Prefix) & \underline{0.00951} & \underline{90.42} & \underline{71.73} \\
 & SK-Tuning (Prompt) & \textbf{0.00838} & \textbf{91.15} & \textbf{71.98} \\ \hline
\multirow{7}{*}{NCBI disease} & Full Fine-tuning & 100.000 & 95.82 & 91.19 \\\cline{2-5}
 & Prefix Tuning & 0.83823 & 82.42 & 63.38 \\
 & Prompt Tuning & 0.13939 & 85.61 & 65.44 \\
 & P-Tuning & 0.15794 & 81.17 & 67.63 \\
 & Lora Rank 2 & 0.13748 & 86.23 & 78.45 \\
 & Lora Rank 4 & 0.71493 & 87.34 & 78.26 \\\cline{2-5}
 & SK-Tuning (Prefix) & \underline{0.00994} & \underline{89.22} & \underline{80.63} \\
 & SK-Tuning (Prompt) & \textbf{0.00438} & \textbf{90.82} & \textbf{81.95} \\ \hline
\multirow{7}{*}{WikiAnn} & Full Fine-tuning & 100.000 & 88.92 & 58.21 \\\cline{2-5}
 & Prefix Tuning & 0.83832 & 74.37 & 31.57 \\
 & Prompt Tuning & 0.01416 & 78.86 & 38.32 \\
 & P-Tuning & 0.15812 & 75.23 & 32.26 \\
 & Lora Rank 2 & 0.13748 & 79.04 & 39.88 \\
 & Lora Rank 4 & 0.71649 & 81.53 & \textbf{44.47} \\\cline{2-5}
 & SK-Tuning (Prefix) & \underline{0.00851} & \underline{82.03} & 42.93 \\
 & SK-Tuning (Prompt) & \textbf{0.00693} & \textbf{83.18} & \underline{43.06} \\ \hline
\end{tabular}%
}
\caption{Token Classification Results for the Phi-2 Model. The best results are highlighted in \textbf{bold}, and the second-best result is \underline{underlined} for clarity except full fine-tuning.}
\label{tab:Token Cls Phi-2}
\end{table}

\subsection{Token Classification}

In this comprehensive analysis of token classification across various datasets, we conducted an extensive evaluation of five distinct models: Bloom, Llama2, Falcon, Mistral, and Phi-2, while exploring a range of fine-tuning techniques to understand their impact on performance, documented in Table \ref{tab:Token Cls Bloom}, \ref{tab:Token Cls Llama2}, \ref{tab:Token Cls Falcon}, \ref{tab:Token Cls Mistral} and \ref{tab:Token Cls Phi-2}. The datasets used for evaluation included conll03, ncbi disease, and wiki ann, each representing different challenges in token classification.

First and foremost, we observed that Full Fine-tuning consistently achieved high accuracy across all models and datasets. However, it also required a substantial percentage of parameters, potentially making it less feasible for resource-constrained environments.

To address the trade-off between model efficiency and performance, we investigated several fine-tuning techniques. Prefix Tuning, Prompt Tuning, and P-Tuning, which involve introducing a small fraction of parameters, showcased mixed results. While these techniques achieved decent accuracy in some cases, they often lagged behind in terms of F1-score, indicating challenges in maintaining a balance between precision and recall.

Remarkably, Lora Rank 2 and Lora Rank 4, with a moderate percentage of parameters, consistently delivered a strong performance, especially in terms of the F1-score. These results underscore the importance of considering the architecture of the model when optimizing for token classification tasks, with Lora Rank models demonstrating their effectiveness.

Finally, SK-Tuning techniques, both Prefix and Prompt variants, stood out as noteworthy approaches. They required an extremely minimal percentage of additional parameters yet yielded competitive accuracy and remarkable F1 scores. This suggests that these techniques have the potential to strike a favorable balance between model efficiency and task effectiveness.





\begin{table}[ht!]
\centering
\resizebox{0.55\textwidth}{!}{%
\begin{tabular}{ccccc}\hline
Dataset & Type & Parameters (\%) & Accuracy (\%) & F1-score (\%) \\\hline
\multirow{8}{*}{RTE} & Full Fine-tuning & 100.000 & 92.31 & 87.19 \\\cline{2-5}
 & Prefix Tuning & 0.03493 & 70.03 & 64.06 \\
 & Prompt Tuning & 0.00714 & 65.34 & 62.20 \\
 & P-Tuning & 0.01584 & 71.11 & 69.23 \\
 & Lora Rank 2 & 0.01402 & 80.25 & 80.01 \\
 & Lora Rank 4 & 0.05804 & \underline{84.45} & \underline{83.26} \\\cline{2-5}
 & SK-Tuning (Prefix) & \underline{0.00076} & 83.88 & 82.76 \\
 & SK-Tuning (Prompt) & \textbf{0.00053} & \textbf{84.92} & \textbf{83.87} \\ \hline
\multirow{8}{*}{MRPC} & Full Fine-tuning & 100.000 & 90.01 & 91.13 \\\cline{2-5}
 & Prefix Tuning & 0.03494 & 73.56 & 81.70 \\
 & Prompt Tuning & 0.00773 & 81.39 & 86.01 \\
 & P-Tuning & 0.01562 & 78.08 & 84.38 \\
 & Lora Rank 2 & 0.01393 & 80.21 & 82.29 \\
 & Lora Rank 4 & 0.05799 & 83.88 & 84.84 \\\cline{2-5}
 & SK-Tuning (Prefix) & \underline{0.00082} & \underline{88.99} & \underline{86.28} \\
 & SK-Tuning (Prompt) & \textbf{0.00054} & \textbf{89.03} & \textbf{86.37} \\ \hline
\multirow{8}{*}{SNLI} & Full Fine-tuning & 100.000 & 95.62 & 95.78 \\\cline{2-5}
 & Prefix Tuning & 0.03492 & 87.32 & 87.26 \\
 & Prompt Tuning & 0.00803 & 88.88 & 88.87 \\
 & P-Tuning & 0.01594 & 86.22 & 86.54 \\
 & Lora Rank 2 & 0.01412 & 91.37 & 91.36 \\
 & Lora Rank 4 & 0.05813 & \underline{93.23} & \underline{93.68} \\\cline{2-5}
 & SK-Tuning (Prefix) & \underline{0.00085} & 92.54 & 92.98 \\
 & SK-Tuning (Prompt) & \textbf{0.00060} & \textbf{93.75} & \textbf{94.02} \\ \hline
\end{tabular}%
}
\caption{Entailment Classification Results for the Bloom Model. The best results are highlighted in \textbf{bold}, and the second-best result is \underline{underlined} for clarity except full fine-tuning.}
\label{tab:Entailment Cls Bloom}
\end{table}

\begin{table}[ht!]
\centering
\resizebox{0.55\textwidth}{!}{%
\begin{tabular}{ccccc}\hline
Dataset & Type & Parameters (\%) & Accuracy (\%) & F1-score (\%) \\\hline
\multirow{8}{*}{RTE} & Full Fine-tuning & 100.000 & 93.51 & 88.92 \\\cline{2-5}
 & Prefix Tuning & 0.03982 & 70.15 & 65.23 \\
 & Prompt Tuning & 0.00737 & 62.81 & 66.00 \\
 & P-Tuning & 0.01753 & 67.24 & 66.21 \\
 & Lora Rank 2 & 0.01612 & 81.04 & 80.67 \\
 & Lora Rank 4 & 0.03224 & \underline{83.43} & 81.44 \\\cline{2-5}
 & SK-Tuning (Prefix) & \underline{0.00077} & \textbf{85.73} & \textbf{84.01} \\
 & SK-Tuning (Prompt) & \textbf{0.00052} & 83.72 & \underline{83.43} \\ \hline
\multirow{8}{*}{MRPC} & Full Fine-tuning & 100.000 & 92.25 & 92.95 \\\cline{2-5}
 & Prefix Tuning & 0.03973 & 79.41 & 80.01 \\
 & Prompt Tuning & 0.00724 & 80.18 & 80.37 \\
 & P-Tuning & 0.01745 & 74.56 & 82.67 \\
 & Lora Rank 2 & 0.01601 & 80.48 & 82.02 \\
 & Lora Rank 4 & 0.03218 & 81.89 & 83.11 \\\cline{2-5}
 & SK-Tuning (Prefix) & \underline{0.00082} & \textbf{85.97} & \textbf{86.37} \\
 & SK-Tuning (Prompt) & \textbf{0.00051} & \underline{85.03} & \underline{85.37} \\ \hline
\multirow{8}{*}{SNLI} & Full Fine-tuning & 100.000 & 93.31 & 94.03 \\\cline{2-5}
 & Prefix Tuning & 0.03986 & 86.34 & 86.33 \\
 & Prompt Tuning & 0.00736 & 87.02 & 87.41 \\
 & P-Tuning & 0.01752 & 85.17 & 86.27 \\
 & Lora Rank 2 & 0.01613 & 90.21 & 90.87 \\
 & Lora Rank 4 & 0.03228 & \underline{91.15} & \underline{91.85} \\\cline{2-5}
 & SK-Tuning (Prefix) & \underline{0.00095} & \textbf{91.43} & \textbf{91.98} \\
 & SK-Tuning (Prompt) & \textbf{0.00069} & 90.97 & 91.04 \\ \hline 
\end{tabular}%
}
\caption{Entailment Classification Results for the Llama2 Model. The best results are highlighted in \textbf{bold}, and the second-best result is \underline{underlined} for clarity except full fine-tuning.}
\label{tab:Entailment Cls Llama2}
\end{table}

\begin{table}[ht!]
\centering
\resizebox{0.55\textwidth}{!}{%
\begin{tabular}{ccccc}\hline
Dataset & Type & Parameters (\%) & Accuracy (\%) & F1-score (\%) \\\hline
\multirow{8}{*}{RTE} & Full Fine-tuning & 100.000 & 93.22 & 87.67 \\\cline{2-5}
 & Prefix Tuning & 0.03822 & 64.23 & 63.38 \\
 & Prompt Tuning & 0.00813 & 66.51 & 66.02 \\
 & P-Tuning & 0.01794 & 53.42 & 53.09 \\
 & Lora Rank 2 & 0.01138 & 73.28 & 70.15 \\
 & Lora Rank 4 & 0.01774 & 78.33 & 73.42 \\\cline{2-5}
 & SK-Tuning (Prefix) & \underline{0.00084} & \underline{80.12} & \underline{79.73} \\
 & SK-Tuning (Prompt) & \textbf{0.00065} & \textbf{80.25} & \textbf{79.78} \\ \hline
MRPC & Full Fine-tuning & 100.000 & 90.21 & 90.83 \\\cline{2-5}
 & Prefix Tuning & 0.03813 & 74.13 & 78.22 \\
 & Prompt Tuning & 0.00715 & 80.04 & 80.19 \\
 & P-Tuning & 0.01783 & 80.43 & 79.59 \\
 & Lora Rank 2 & 0.00983 & 80.82 & 82.21 \\
 & Lora Rank 4 & \underline{0.01763} & 82.52 & 83.01 \\\cline{2-5}
 & SK-Tuning (Prefix) & 0.00079 & 82.68 & 83.68 \\ 
 & SK-Tuning (Prompt) & \textbf{0.00054} & 83.03 & 85.37 \\\hline
SNLI & Full Fine-tuning & 100.000 & 92.53 & 92.97 \\\cline{2-5}
 & Prefix Tuning & 0.03822 & 84.33 & 84.98 \\
 & Prompt Tuning & 0.00843 & 86.13 & 86.93 \\
 & P-Tuning & 0.01782 & 83.31 & 83.66 \\
 & Lora Rank 2 & 0.01163 & 87.05 & 87.29 \\
 & Lora Rank 4 & 0.06773 & 89.21 & 89.88 \\\cline{2-5}
 & SK-Tuning (Prefix) & \underline{0.00072} & \textbf{90.86} & \underline{91.00} \\
 & SK-Tuning (Prompt) & \textbf{0.00053} & \textbf{90.86} & \underline{91.00} \\ \hline
\end{tabular}%
}
\caption{Entailment Classification Results for the Falcon Model. The best results are highlighted in \textbf{bold}, and the second-best result is \underline{underlined} for clarity except full fine-tuning.}
\label{tab:Entailment Cls Falcon}
\end{table}

\begin{table}[ht!]
\centering
\resizebox{0.55\textwidth}{!}{%
\begin{tabular}{ccccc}\hline
Dataset & Type & Parameters (\%) & Accuracy (\%) & F1-score (\%) \\\hline
\multirow{8}{*}{RTE} & Full Fine-tuning & 100.000 & 94.67 & 89.82 \\\cline{2-5}
 & Prefix Tuning & 0.03663 & 76.22 & 74.45 \\
 & Prompt Tuning & 0.00732 & 80.34 & 80.17 \\
 & P-Tuning & 0.01778 & 75.12 & 75.86 \\
 & Lora Rank 2 & 0.01521 & 83.39 & 82.25 \\
 & Lora Rank 4 & 0.06739 & \underline{85.65} & 83.12 \\\cline{2-5}
 & SK-Tuning (Prefix) & \underline{0.00083} & 84.73 & \underline{83.87} \\
 & SK-Tuning (Prompt) & \textbf{0.00065} & \textbf{85.94} & \textbf{84.67} \\ \hline
\multirow{8}{*}{MRPC} & Full Fine-tuning & 100.000 & 93.02 & 94.21 \\\cline{2-5}
 & Prefix Tuning & 0.03654 & 75.28 & 77.03 \\
 & Prompt Tuning & 0.00722 & 80.34 & 82.17 \\
 & P-Tuning & 0.01715 & 76.19 & 80.31 \\
 & Lora Rank 2 & 0.01513 & 82.83 & 83.41 \\
 & Lora Rank 4 & 0.06724 & \textbf{86.47} & \underline{87.02} \\\cline{2-5}
 & SK-Tuning (Prefix) & \underline{0.00082} & 85.63 & 85.17 \\
 & SK-Tuning (Prompt) & \textbf{0.00055} & \underline{86.31} & \textbf{87.98} \\ \hline
\multirow{8}{*}{SNLI} & Full Fine-tuning & 100.000 & 94.21 & 95.32 \\\cline{2-5}
 & Prefix Tuning & 0.03666 & 85.55 & 85.78 \\
 & Prompt Tuning & 0.00744 & 86.35 & 86.21 \\
 & P-Tuning & 0.01774 & 85.37 & 86.05 \\
 & Lora Rank 2 & 0.01524 & 84.12 & 84.76 \\
 & Lora Rank 4 & 0.06736 & 89.11 & 89.77 \\\cline{2-5}
 & SK-Tuning (Prefix) & \underline{0.00089} & \underline{91.62} & \underline{91.31} \\
 & SK-Tuning (Prompt) & \textbf{0.00068} & \textbf{92.56} & \textbf{91.86} \\\hline
\end{tabular}%
}
\caption{Entailment Classification Results for the Mistral Model. The best results are highlighted in \textbf{bold}, and the second-best result is \underline{underlined} for clarity except full fine-tuning.}
\label{tab:Entailment Cls Mistral}
\end{table}

\begin{table}[ht!]
\centering
\resizebox{0.55\textwidth}{!}{%
\begin{tabular}{ccccc}\hline
Dataset & Type & Parameters (\%) & Accuracy (\%) & F1-score (\%) \\\hline
\multirow{8}{*}{RTE} & Full Fine-tuning & 100.000 & 90.37 & 85.74 \\\cline{2-5}
 & Prefix Tuning & 0.83872 & 59.54 & 58.27 \\
 & Prompt Tuning & 0.14234 & 61.18 & 61.84 \\
 & P-Tuning & 0.15834 & 58.61 & 56.38 \\
 & Lora Rank 2 & 0.13746 & 66.52 & 65.82 \\
 & Lora Rank 4 & 0.71658 & 72.25 & 70.45 \\\cline{2-5}
 & SK-Tuning (Prefix) & \underline{0.00424} & \underline{76.44} & \underline{75.92} \\
 & SK-Tuning (Prompt) & \textbf{0.00252} & \textbf{76.53} & \textbf{76.11} \\ \hline
\multirow{8}{*}{MRPC} & Full Fine-tuning & 100.000 & 89.31 & 90.21 \\\cline{2-5}
 & Prefix Tuning & 0.83822 & 71.15 & 72.78 \\
 & Prompt Tuning & 0.14345 & 73.16 & 75.28 \\
 & P-Tuning & 0.15842 & 70.48 & 71.21 \\
 & Lora Rank 2 & 0.13747 & 80.53 & 81.33 \\
 & Lora Rank 4 & 0.71659 & \underline{83.19} & \underline{84.23} \\\cline{2-5}
 & SK-Tuning (Prefix) & \underline{0.00742} & \textbf{83.63} & \textbf{84.72} \\
 & SK-Tuning (Prompt) & \textbf{0.00349} & 82.54 & 83.42 \\ \hline
\multirow{8}{*}{SNLI} & Full Fine-tuning & 100.00 & 90.54 & 91.02 \\\cline{2-5}
 & Prefix Tuning & 0.83844 & 79.27 & 79.82 \\
 & Prompt Tuning & 0.14149 & 81.30 & 81.80 \\
 & P-Tuning & 0.15823 & 78.56 & 77.96 \\
 & Lora Rank 2 & 0.13745 & 82.45 & 82.67 \\
 & Lora Rank 4 & 0.71656 & 84.36 & 84.89 \\\cline{2-5}
 & SK-Tuning (Prefix) & \underline{0.00609} & \textbf{89.21} & \textbf{90.51} \\
 & SK-Tuning (Prompt) & \textbf{0.00589} & \underline{88.62} & \underline{88.95} \\\hline
\end{tabular}%
}
\caption{Entailment Classification Results for the Phi-2 Model. The best results are highlighted in \textbf{bold}, and the second-best result is \underline{underlined} for clarity except full fine-tuning.}
\label{tab:Entailment Cls Phi-2}
\end{table}

\subsection{Entailment Detection}
The results of entailment detection using various models, including Bloom, Llama2, Falcon, Mistral, and Phi-2, are presented in Table \ref{tab:Entailment Cls Bloom}, \ref{tab:Entailment Cls Llama2}, \ref{tab:Entailment Cls Falcon}, \ref{tab:Entailment Cls Mistral} and \ref{tab:Entailment Cls Phi-2} . Across all three datasets (RTE, MRPC, SNLI), full fine-tuning consistently achieves the highest accuracy and F1-score, with Bloom and Mistral models demonstrating remarkable results. This underscores the value of fine-tuning the entire model's parameters to adapt to specific entailment tasks, as it allows the model to capture intricate patterns and nuances in the data.

In contrast, prefix tuning and prompt tuning techniques, which involve fine-tuning only a small fraction of the model's parameters, tend to yield significantly lower accuracy and F1-scores. This suggests that limiting parameter updates to specific prefixes or prompts may not be sufficient for optimal entailment classification performance, as these methods may struggle to capture the diverse and complex relationships present in the data.

The Lora Rank 2 and Lora Rank 4 models deliver competitive results, particularly evident in the RTE dataset, where they outperform other techniques. This indicates that techniques like Lora Rank, which involve a moderate amount of parameter modification, can strike a balance between model adaptation and computational efficiency.

However, SK-Tuning, whether applied to prefixes or prompts, consistently performs well across datasets, demonstrating its effectiveness as an alternative fine-tuning strategy. SK-Tuning achieves strong results with a minimal increase in the number of parameters, making it a promising approach for entailment classification tasks where computational resources are a concern.




\section{Ablation Study}
\label{ablation_study}
\subsection{Efficiency}

\begin{figure}[ht!]
    \centering
    \begin{minipage}[t]{.470\textwidth}
        \centering
        \includegraphics[width=0.999\linewidth]{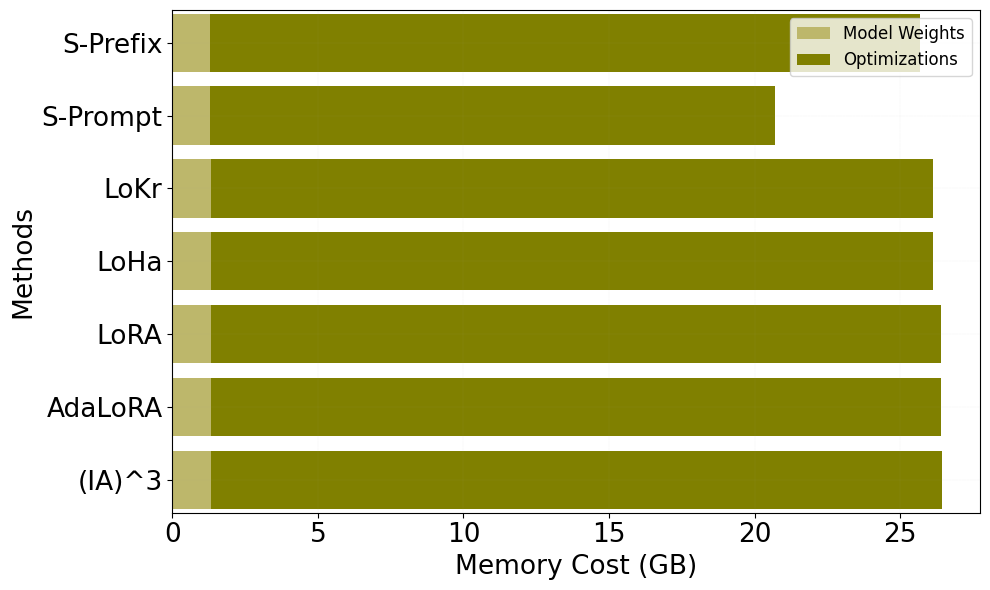}
  
    \end{minipage}%
    \hfill
    \begin{minipage}[t]{.530\textwidth}
        \centering
        \includegraphics[width=0.999\linewidth]{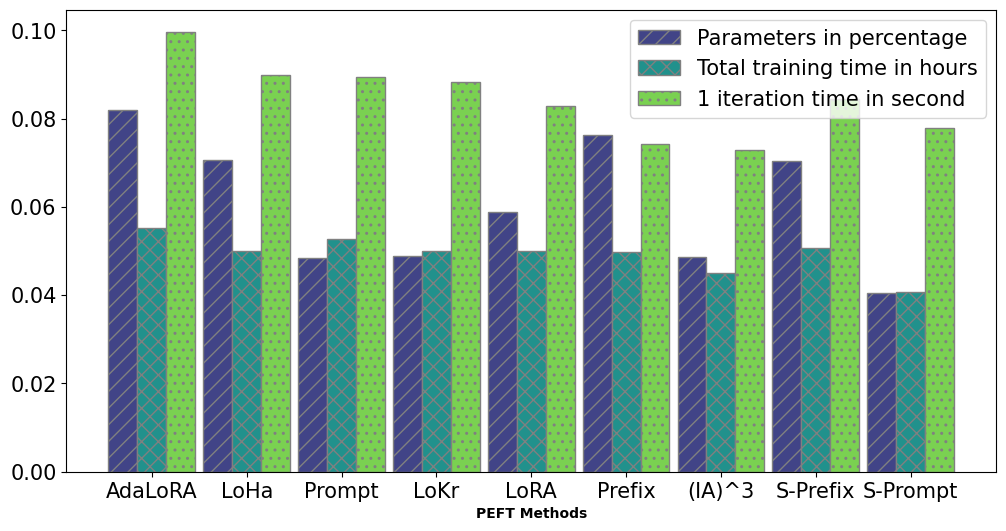}

    \end{minipage}
    \caption{Comparison of memory efficiency (left) and training efficiency (right) across various PEFT methods. S-Prefix and S-Prompt represent SK-Tuning applied to prefix tuning and prompt tuning, respectively. The left chart shows the memory cost in GB, highlighting the model weights and optimizations, while the right chart displays the percentage of parameters, total training time in hours, and iteration time per second.} \label{fig:efficiancy}
\end{figure}

Figure \ref{fig:efficiancy} illustrates that SK-Tuning methods for Prefix and Prompt, demonstrate superior memory efficiency with the lowest memory cost among the compared PEFT methods, making them ideal for resource-constrained environments. Despite their minimal memory footprint, these methods maintain competitive training efficiency, balancing low parameter percentages with moderate training times, which highlights their effectiveness in achieving lightweight and fast fine-tuning. Compared to other methods like LoKr, LoHa, and LoRA, which show higher memory costs and varying degrees of training efficiency, SK-Tuning stands out as a robust approach that optimizes both memory and computational resources, making it particularly advantageous for scenarios where efficiency is paramount.
\subsection{Faster Convergence with SK-Tuning}

In this section, we present an ablation study comparing the convergence speed and performance of SK-Tuning with traditional prompt and prefix tuning methods on three different downstream tasks: Token Classification, Sequence Classification, and NLI. We hypothesize that SK-Tuning, leveraging semantic knowledge, will lead to faster convergence due to the inherent zero-shot capabilities of LLMs  \cite{kojima2022large}.

\begin{figure}[ht!]
    \centering
    \begin{minipage}[t]{.50\textwidth}
        \centering
        \includegraphics[width=0.999\linewidth]{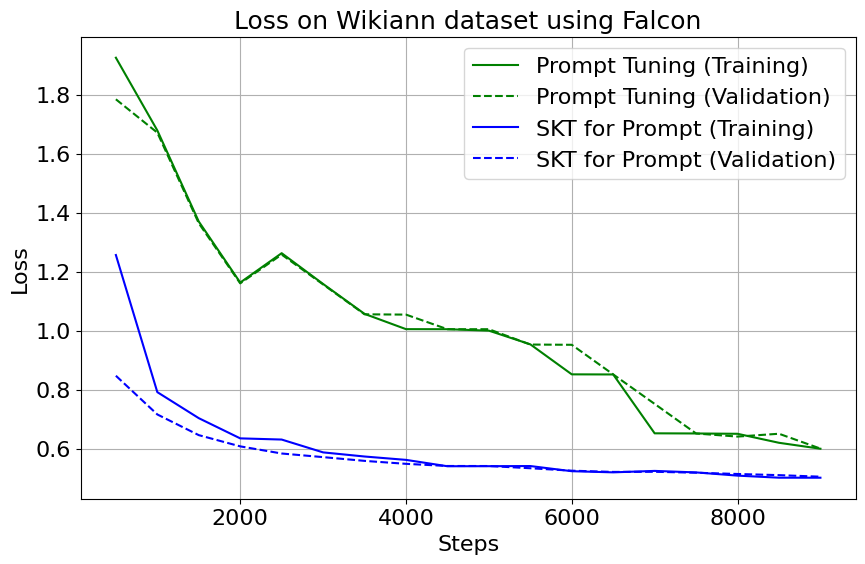}
  
    \end{minipage}%
    \hfill
    \begin{minipage}[t]{.50\textwidth}
        \centering
        \includegraphics[width=0.999\linewidth]{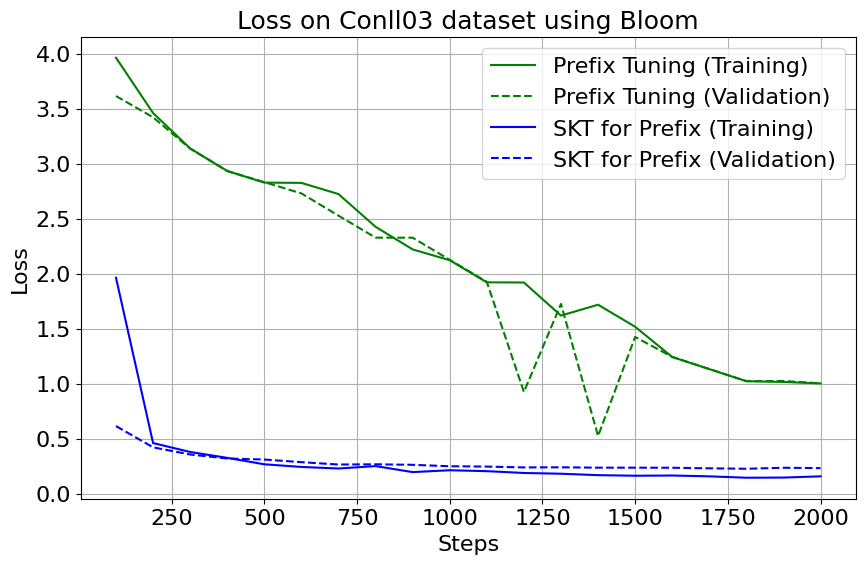}

    \end{minipage}
    \caption{Convergence Comparison for Token Classification}
    \label{fig:token_classification}
\end{figure}

\subsubsection{Accelerated Convergence in Token Classification}

In the context of token classification tasks, we conducted a comprehensive comparison between SK-Tuning and traditional tuning methods. We utilized two benchmark datasets, namely Wikiann and Conll03, both featuring token-level labels. Our primary objective was to analyze the convergence behavior, measured in terms of loss reduction, as training steps progressed.

Figure~\ref{fig:token_classification} visually presents the convergence trajectories for SK-Tuning and traditional methods. Notably, we observed a remarkable disparity in the convergence speed between these approaches. SK-Tuning, whether applied to prefixes or prompts, demonstrated a strikingly swift convergence compared to the conventional tuning method.

This accelerated convergence showcased in Figure~\ref{fig:token_classification} serves as compelling evidence of the significant advantages brought about by the incorporation of semantic knowledge. It underscores the ability of SK-Tuning to facilitate rapid adaptation to the intricacies of token classification tasks, emphasizing the practical utility of this approach.

\begin{figure*}[ht!]
    \centering
    \begin{minipage}[t]{.50\textwidth}
        \centering
        \includegraphics[width=0.999\linewidth]{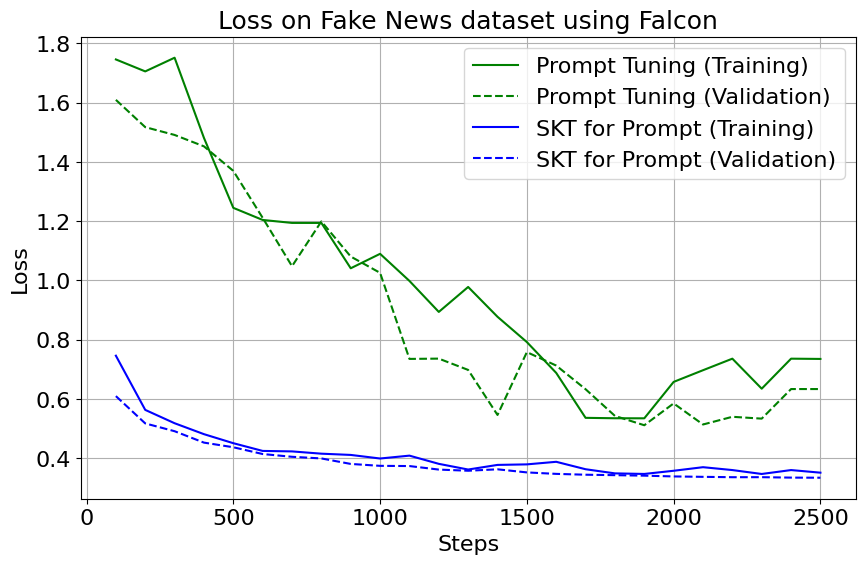}
  
    \end{minipage}%
    \hfill
    \begin{minipage}[t]{.50\textwidth}
        \centering
        \includegraphics[width=0.999\linewidth]{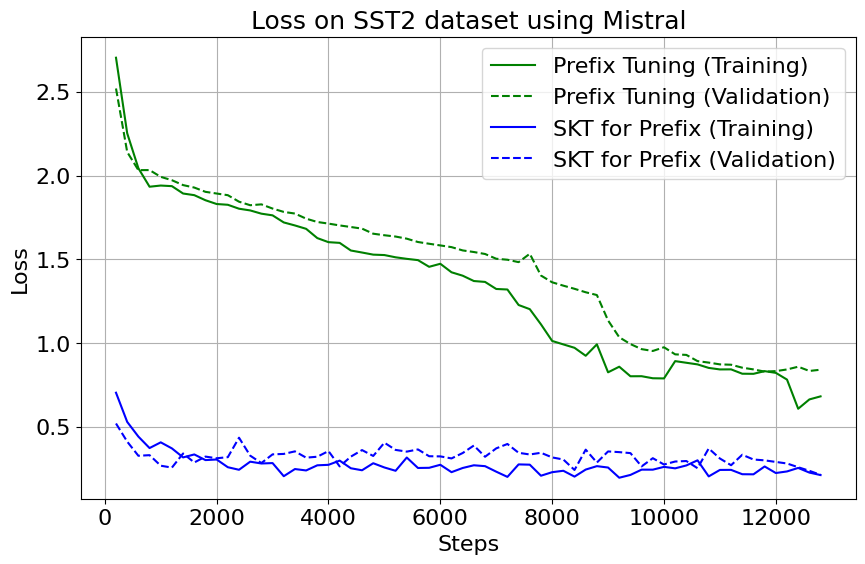}

    \end{minipage}
    \caption{Convergence Comparison for Sequence Classification}
    \label{fig:sequence_classification}
\end{figure*}

\subsubsection{Accelerated Convergence in Sequence Classification}

For the evaluation of SK-Tuning in sequence classification tasks, we conducted a comparative analysis against traditional tuning methods. Our experimentation leveraged two benchmark datasets: Fake News and SST2, both featuring sequences with corresponding labels. Our primary objective was to assess the convergence performance, measured in terms of loss reduction, as the model underwent training iterations.

Figure~\ref{fig:sequence_classification} offers a visual representation of the convergence patterns observed during sequence classification. Notably, the results depicted in the figure demonstrate the accelerated convergence achieved with SK-Tuning when compared to conventional tuning methods.

The swift convergence illustrated in Figure~\ref{fig:sequence_classification} underscores the significant advantages bestowed by the integration of semantic knowledge into the fine-tuning process. This enhancement enables the model to quickly adapt to the nuances of the specific sequence classification task, reaffirming the effectiveness of SK-Tuning in practical scenarios.

\begin{figure*}[!t]
    \centering
    \begin{minipage}[t]{.50\textwidth}
        \centering
        \includegraphics[width=0.999\linewidth]{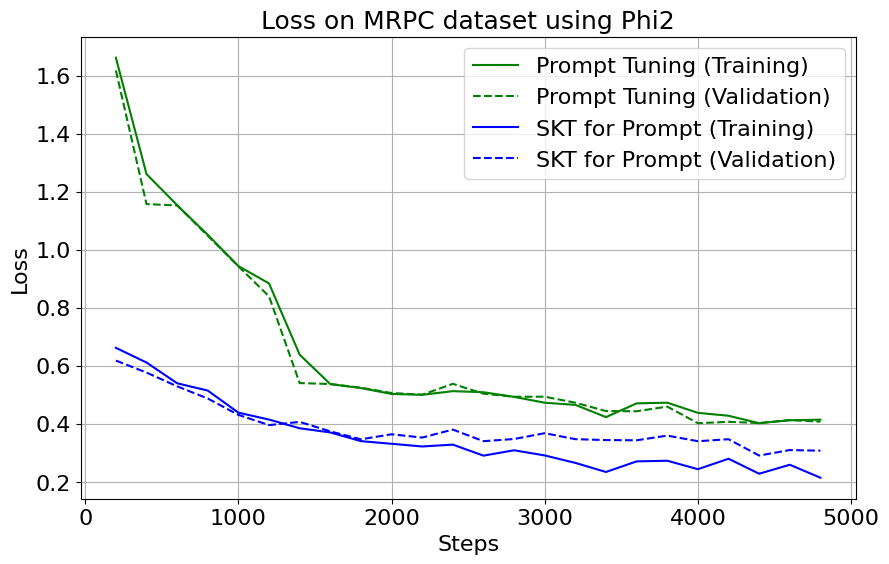}
  
        \label{fig:sub1}
    \end{minipage}%
    \hfill
    \begin{minipage}[t]{.50\textwidth}
        \centering
        \includegraphics[width=0.999\linewidth]{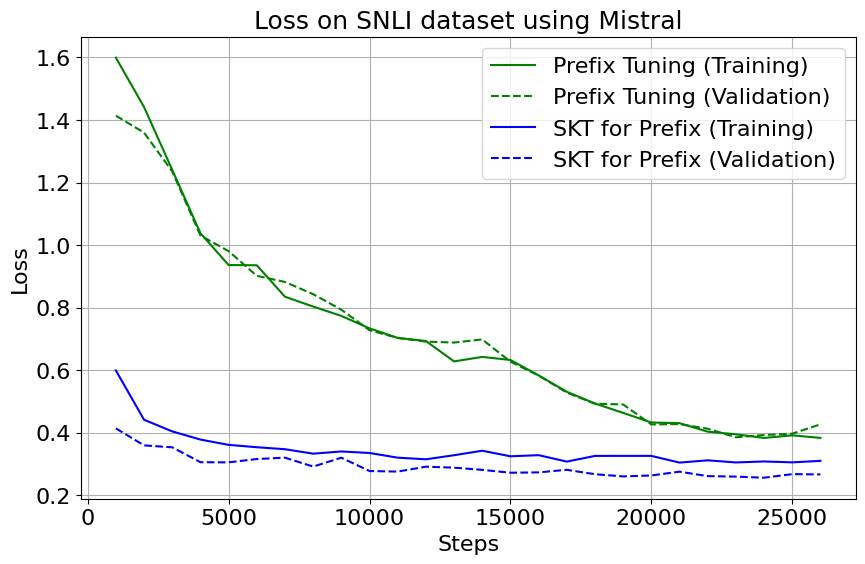}

    \end{minipage}
    \caption{Convergence Comparison for Sequence Classification}
    \label{fig:nli}
\end{figure*}

\subsubsection{Accelerated Convergence in NLI}

In the realm of NLI tasks, we conducted a comparative analysis pitting SK-Tuning against traditional tuning methods. Our evaluation incorporated well-established datasets, including MRPC and SNLI, which consist of premise-hypothesis pairs and their corresponding entailment labels. The primary objective was to assess convergence speed, measured in terms of training steps.

Figure~\ref{fig:nli} visually illustrates the convergence dynamics observed during NLI tasks. Notably, the findings showcased in the figure reveal the expedited convergence achieved through SK-Tuning when compared to traditional tuning approaches.

The swift convergence depicted in Figure~\ref{fig:nli} underscores the substantial advantages conferred by the integration of semantic knowledge into the fine-tuning process. This augmentation enhances the model's adaptability, enabling it to quickly grasp the nuances of NLI tasks and reaffirming the practical utility of SK-Tuning in advancing NLI model performance.

Our ablation study clearly demonstrates that SK-Tuning outperforms traditional prompt and prefix tuning methods in terms of convergence speed across a range of downstream tasks. The incorporation of semantic knowledge, along with the zero-shot capabilities of LLMs, contributes to faster task adaptation. Additionally, SK-Tuning consistently leads to better performance, as shown in subsequent sections.

\subsection{Adapter Layers}
In this study, we investigate the impact of adapter layer complexity on the performance of fine-tuned models. Specifically, we analyze how increasing the complexity of adapter layers affects various factors, including the percentage of parameters, computational cost, and convergence speed. We conducted experiments using the Mistral 7B model on the SST2 dataset, and the results are presented in Table \ref{tab:adapter_test}.

\begin{table}[ht]
\centering
\resizebox{0.65\textwidth}{!}{%
\begin{tabular}{cccccc}\hline
 \ Number of Layers & Parameters & Accuracy (\%) & F1-score (\%)  & Convergence Steps \\\hline
    1 & 0.00031 & 96.93 & 97.14 & 1500 \\ \hline
    3 & 0.02236 & 95.32 & 97.02 & 7900 \\ \hline
    5 & 1.18583 & 96.98 & 97.25 & 16600 \\ \hline
    7 & 3.46256 & 97.21 & 97.54 & 23900 \\ \hline
    9 & 6.28583 & 97.16 & 97.37 & 53200 \\ \hline
    11 & 9.89372 & 97.29 & 97.87 & 79400 \\ \hline
    
\end{tabular}
}
\caption{Exploring the Trade-offs - Adapter Complexity vs. Performance}
\label{tab:adapter_test}
\end{table}

As shown in Table \ref{tab:adapter_test}, increasing the number of adapter layers leads to a proportional increase in the number of parameters. This rise in complexity comes at the cost of increased computational resources and slower convergence. While the performance of the model does show marginal improvements with more complex adapter layers, it is essential to note that these gains are relatively modest.

For instance, with just one adapter layer, the model exhibits a relatively small number of parameters, efficient convergence, and high accuracy. However, as we progressively increase the complexity with additional layers, the number of parameters surges significantly, computational requirements escalate, and convergence becomes substantially slower. Notably, the performance gains achieved by complex adapter layers are relatively modest.

The observed trend suggested that as the complexity of the adapter layers increased, the computational demands and training time also increased substantially. This phenomenon can be attributed to the need for extensive training to capture and leverage semantic information effectively.

\subsection{Effect of Prompt and Prefix Text}
\label{sec:effect_prompt}
In this ablation study, we investigate the influence of prompt and prefix text length on the performance of SK-Tuning for sentiment classification using the SST-2 dataset. Our goal is to demonstrate that well-crafted prompt or prefix texts can outperform longer, less informative alternatives, despite the latter offering a larger number of trainable parameters.

We conducted experiments with various prompt and prefix texts and evaluated their corresponding accuracy on sentiment classification tasks using the Mistral model, which boasts 7 billion parameters. The table below summarizes the results:

\begin{table}[ht]
\centering
\resizebox{0.75\textwidth}{!}{%
\begin{tabular}{p{11cm}|c|c|c}
\hline
\textbf{Prompt / Prefix (Text)} & \textbf{Prefix} & \textbf{Prompt} & \textbf{Token Length} \\
\hline
"Classify the sentiment of the following text" & 93.41 & 92.83 & 9 \\
\hline
"I have a piece of text and I need to understand its emotional tone. Could you classify the sentiment of the text:" & \underline{96.04} & \underline{95.41} & 29 \\
\hline
"Considering the context and tone, can you classify the sentiment of the following text? Here's the text" & 95.13 & 94.82 & 24 \\
\hline
"Focus on the emotional cues present in the text:" & 92.93 & 93.09 & 12 \\
\hline
"Let's analyze the sentiment of this text together. I'll provide the text, and you classify the sentiment of the text. Here's the text:" & 94.78 & 94.17 & 35 \\
\hline
"Classify the positive or negative sentiment of the text:" & \textbf{96.83} & \textbf{96.52} & 11 \\
\hline
\end{tabular}
}
\caption{Effect of Prompt and Prefix Length on Sentiment Classification Accuracy}
\label{tab:prompt-prefix-length}
\end{table}

The results presented in Table~\ref{tab:prompt-prefix-length} clearly illustrate that a concise and informative prompt text outperforms longer and less focused alternatives. Despite the fact that longer prompts or prefixes provide more trainable parameters, our findings underscore the significance of crafting prompts that offer clear task instructions and context, resulting in enhanced model performance.

\begin{figure*}[!t]
  \centering
  \includegraphics[width=\linewidth]{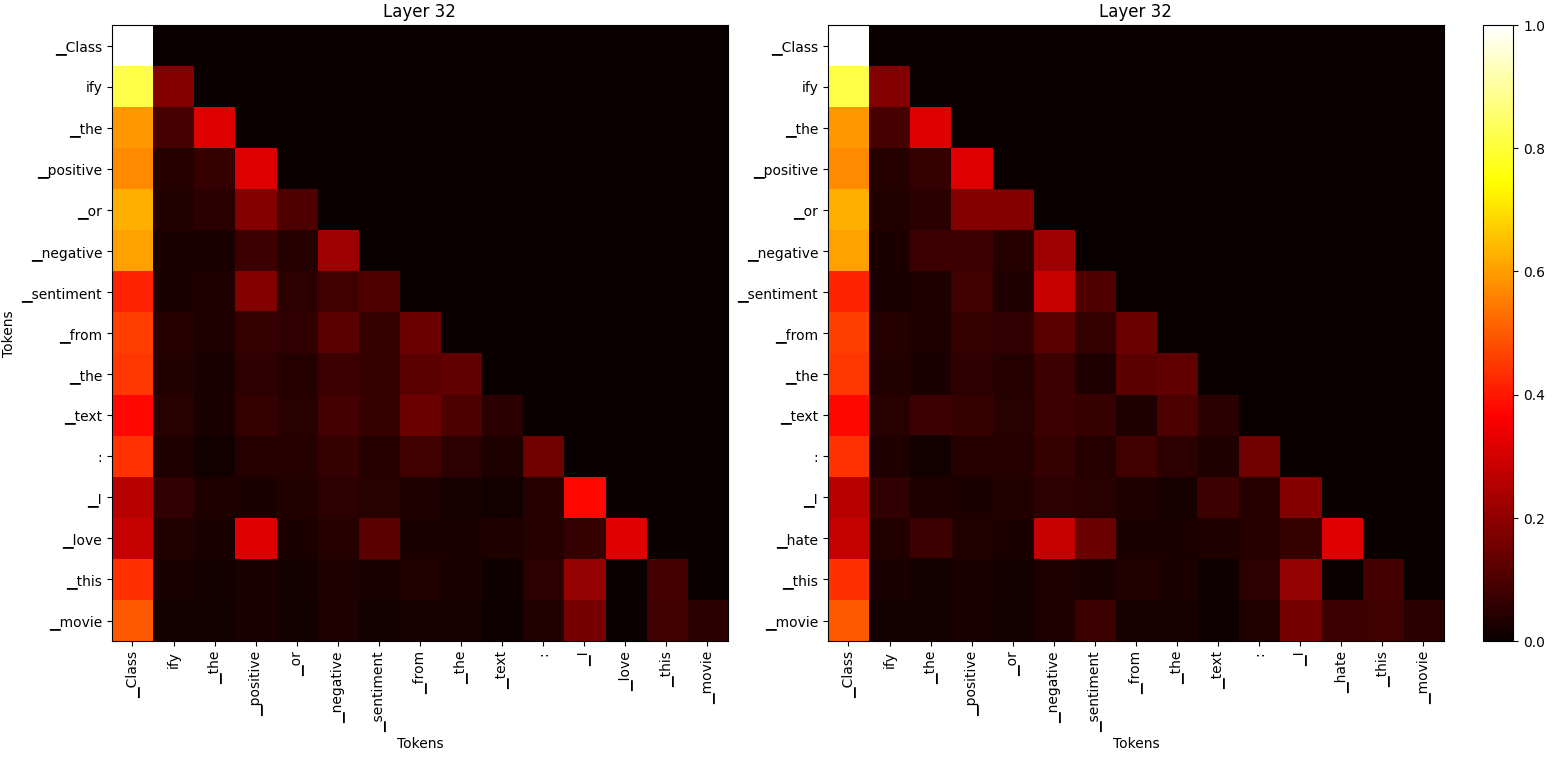}
    \caption{ Attentional Insights - Exploring the Sentimental Connection between Prompt Text and Input Text. The left side of the figure reveals the attention scores between the prompt text, particularly the word `positive,' and the input text 'I love this movie.' On the right side, the attention scores depict the relationship between the prompt word 'negative' and the input text 'I hate this movie.' These attention patterns shed light on how well-crafted prompt texts enhance the model's decision-making process.}
  \label{fig:pos_neg}
\end{figure*}
Furthermore, to visualize the relationship between the prompt text and the input text, we analyzed the attention scores of the last layer. Specifically, we used the prompt text \textit{\textcolor{red}{Classify the positive or negative sentiment of the text.}} in conjunction with input texts \textit{ \textcolor{blue}{I love this movie.}} and \textit{\textcolor{blue}{I hate this movie.}} The figures in Figure~\ref{fig:pos_neg} depict the attention scores, highlighting the sentimental connection between the prompt and the text. In the left figure, the prompt text, particularly the word \textit{\textcolor{red}{positive}} exhibits a strong attention score with  \textit{\textcolor{blue}{love}} Conversely, in the right figure, the prompt word \textit{\textcolor{red}{negative}} shows a pronounced attention score with \textit{\textcolor{blue}{hate}} This observation suggests that including words like \textit{\textcolor{red}{positive}} and \textit{\textcolor{red}{negative}} in the prompt text significantly aids the model in making informed decisions, thereby emphasizing the importance of crafting effective prompt texts.

\section{Discussion}
\label{discussion}
We present a comprehensive comparison of our proposed SK-Tuning method with established parameter-efficient fine-tuning techniques, including Prompt Tuning, Prefix Tuning, P-Tuning, LORA Rank 2, and LORA Rank 4. Our evaluation encompassed a diverse set of downstream tasks across various domains within NLP. Notably, SK-Tuning for both prompts and prefixes consistently outperformed these traditional methods across several key metrics, including accuracy, F1 score, and parameter efficiency.

One of the key takeaways from our comparison is the remarkable performance gains achieved by SK-Tuning. In terms of accuracy and F1 score, SK-Tuning consistently delivered superior results across the spectrum of tasks. This improvement underscores the effectiveness of leveraging semantically meaningful information in the fine-tuning process, as opposed to relying on arbitrary virtual tokens.

Equally noteworthy is the efficiency of SK-Tuning. By minimizing the number of trainable parameters required for adaptation, our approach demonstrates a substantial reduction in computational resources while maintaining or even enhancing task performance. This efficiency is particularly crucial in practical applications, where resource constraints often play a significant role.

Another noteworthy aspect of our study is the extensive evaluation across five different pretrained LLMs: Bloom (7B), Falcon (7B), LLAMA2 (7B), Mistral (7B), and Phi2 (2.7B). Our results consistently indicate that SK-Tuning is a robust and versatile technique that can be applied to various LLM architectures, demonstrating its broad applicability and effectiveness across different model sizes and complexities.

\section{Limitations}
\label{limitations}

While SK-Tuning offers significant advantages in terms of performance and parameter efficiency, there are several key limitations that should be considered:

\subsection{Training and Inference Time Overhead}
One of the primary limitations of SK-Tuning is the potential increase in inference or training time. Since it utilizes the pre-trained LLM twice during the forward pass: once to obtain semantic information from the prompt or prefix and again for processing the input data to get output. This dual usage of the LLM can lead to longer training and inference time.

\subsection{Dependency on Pretrained Models}
 SK-Tuning relies heavily on the quality and capabilities of the underlying pretrained LLM. The success of prompt or prefix text tuning is linked to the zero-shot capabilities of the LLM. If the pretrained model does not have a strong grasp of semantic knowledge or lacks certain linguistic skills, the effectiveness of SK-Tuning could be reduced. It needs significant training to accurately understand the semantic meaning of the prompt or prefix text.
\subsection{Semantic Knowledge Acquisition}
The effectiveness of SK-Tuning depends on using prompts or prefixes that are meaningful. The more relevant the prompt is to the task, the better the performance, described in Section \ref{sec:effect_prompt}. However, creating or finding these meaningful prompts can be difficult and might require specific knowledge about the domain. This challenge could limit how useful SK-Tuning is for certain tasks or datasets.

\subsection{Tuning Hyperparameters}
Like other fine-tuning approaches, SK-Tuning involves hyperparameter tuning, including the design of the adapter architecture, the choice of semantic knowledge text, and the adjustment of task-specific modules. Identifying the optimal hyperparameters can be a time-consuming and computationally intensive process.

\section{Conclusion}
\label{conclusion}

In conclusion, our work introduces SK-Tuning as a pioneering approach to fine-tuning LLMs for specific downstream tasks, with a strong emphasis on parameter efficiency. We have shown that traditional methods, relying on learnable virtual tokens in adapters while keeping the LLM's core parameters frozen, often fall short in terms of both efficiency and performance.

SK-Tuning, on the other hand, revolutionizes the fine-tuning process by replacing arbitrary virtual tokens with real, semantically meaningful prefixes. This innovation allows LLMs to tap into their intrinsic semantic knowledge, significantly reducing the need for extensive training iterations. Our experimental results across a range of downstream tasks, including sequence classification, token classification, and NLI, provide compelling evidence that SK-Tuning outperforms traditional approaches. Notably, this improvement is achieved with a reduced number of trainable parameters, emphasizing the efficiency of our method.

By prioritizing parameter efficiency and harnessing the latent semantic understanding of LLMs, SK-Tuning opens up new possibilities for efficient model adaptation across various real-world applications. We believe that our approach holds great promise for advancing the field of NLP, offering researchers and practitioners a valuable tool for achieving enhanced task performance while optimizing computational resources. As LLMs continue to play a pivotal role in NLP, SK-Tuning represents a significant step forward in harnessing their full potential.

\section*{Author contributions statement}

Conceptualization, M.K.; Methodology, N.J,  M.K., A.M., and M.S.I.S; software,  N.J, M.K., A.M., M.S.I.S, and N.J; formal analysis, M.K, P.B; investigation, A.M., M.S.I.S, M.K. and N.J. resources P.B, J.X. and A.M.,  data collection M.S.I.S, A.M, and M.K, writing original draft preparation M.K., A.M., M.S.I.S, and N.J; writing review and editing, P.B, N.Y, O.O.G, and M.K.; visualization M.K., A.M., M.S.I.S, and N.J; supervision, N.Y, O.O.G, and P.B. All authors reviewed the manuscript.

\section*{Data Availability Statement}
Data information and URL are provided within the manuscript.

\bibliography{sample}







 








\end{document}